\documentclass[10pt]{article}
\usepackage[utf8]{inputenc}
\usepackage{amsmath}
\usepackage{amsfonts}
\usepackage{amssymb}
\usepackage{graphicx}
\usepackage{hyperref}
\usepackage{cleveref}
\usepackage{caption}
\usepackage{multirow}
\usepackage{rotating}
\usepackage[width=122mm,left=12mm,paperwidth=146mm,height=193mm,top=12mm,paperheight=217mm]{}
\usepackage{authblk}
\usepackage{fancyhdr}

\begin{document}

\pagestyle{fancy}
\rhead{}
\lhead{U-CATCH}

\title{U-CATCH: Using Color ATtribute of image patCHes in binary descriptors}

\author[1]{Ozgur Yilmaz}
\author[2]{Alisher Abdulkhaev}
\affil[1,2]{Turgut \"{O}zal University, Department of Computer Engineering, Ankara Turkey}

\maketitle
\begin{center}
\noindent\rule{10cm}{0.5pt}
\end{center}

\begin{abstract}
In this study, we propose a simple yet very effective method for extracting color information through binary feature description framework. Our method expands the dimension of binary comparisons into RGB and YCbCr spaces, showing more than 100\% matching improvement compared to non-color binary descriptors for a wide range of hard-to-match cases. The proposed method is general and can be applied to any binary descriptor to make it color sensitive. It is faster than classical binary descriptors for RGB sampling due to the abandonment of grayscale conversion and has almost identical complexity (insignificant  compared to smoothing operation) for YCbCr sampling.
\\[0.5cm]
\end{abstract}

\begin{center}
\noindent\rule{10cm}{0.5pt}
\end{center}

\section{Introduction}
Local feature descriptors are widely used in applications such as 3D reconstruction \cite{reference7}, object detection\cite{reference6}, scene recognition\cite{lazebnik2006beyond}. SIFT\cite{reference9}, HOG \cite{reference10}, SURF\cite{reference11}, GLOH\cite{reference12} use histogram of image gradients and/or gradient orientations, and they have been very successfully applied to many problems in computer vision that require correspondence matching.

A taxonomy of descriptors (BRIEF, ORB, BRISK, LDAHash, DAISY \cite{tola2010daisy}, SIFT, SURF, etc) based on their computational and storage requirements are given in \cite{reference13}. Heinly et al., emphasize that binary descriptors are faster to compute and match also requiring lower storage compared to non-binary descriptors. 
\par
Binary descriptors \cite{reference13} provide a great advantage among feature descriptors in terms of speed without losing too much accuracy. The reason is twofold: 1. relative intensity captures most of the information in image regions for matching \cite{reference19}, 2. similarity evaluation of two binary vectors can be done much faster compared to real-valued vectors of non-binary descriptors. This similarity (Hamming distance) of two binary vectors can be performed by counting the incompatible bits via XOR operation and XOR can be done very fast on modern CPUs \cite{reference2}.
\par
Binary descriptors compare the pixel or average patch values according to a spatial sampling pattern and generate a binary vector. They exploit a fixed sampling pattern but the generation of the pattern varies: BRIEF (Binary Robust Independent Elementary Features) uses a random sampling pattern, while LATCH (Learned Arrangements of Three Patch Codes) exploits learned patch triplet arrangements. 
\par 
Although, BRIEF and LATCH descriptors perform well at feature matching, there is no utilization of color attribute, which is a very important property of visual objects. Binary descriptors transform the input images into gray scale. Since modern cameras are RGB, classical binary descriptors throw away information that might be valuable for discrimination. In literature, there are some studies \cite{reference4} which use color channels independently on existing feature extraction algorithms. For example RGB-SURF algorithm computes features from each color channel and then combine them to get the overall color representation. In this study we propose a new perspective for exploiting color information in images using binary descriptors. We sampled the pixel locations in three dimensions, without treating each channel separately or dividing the image into color channels. We show that the cross-channel comparison via 3D sampling captures the color property of image patches very effectively and efficiently. This approach can be used in any binary descriptor regardless of their sampling patterns. In this paper, we applied our idea on BRIEF and LATCH descriptors. In our experiments, we sampled pixel locations in two color spaces: RGB and YCbCr. Since RGB and YCbCr color spaces represents chromaticity in distinct ways, we experimented with both of them for completeness and show 100\% improvement in matching performance. 

\section{Review of Binary and Color Descriptors} \label{section2}

\subsection{BRIEF}
BRIEF\cite{reference2} was the first proposed binary descriptor with surprisingly well matching performance. It suggests that for discrimination among feature locations in the image, relative intensity holds most of the information, which is also exploited in local binary pattern approaches \cite{reference19}. 
It samples the pixel locations around a key point randomly due to a fixed distribution (Equation \ref{eq:distribution_BRIEF}). However it has been shown that non-random, i.e. learned sampling pattern can be superior \cite{reference14}. BRIEF compares two pixel intensities in order to build a binary vector. Binary comparison of intensity values is very noise-sensitive especially around relatively uniform regions. In order to make the descriptor less sensitive to noise, smoothing is applied on the window (size 48x48). Gaussian smoothing is used, with the standart deviation of 2 and kernel window size of size 9x9. Effect of pre-smoothing is very significant as stated in \cite{reference2}.

\par
Spatial arrangements of binary tests are randomly sampled from the probability distribution function shown in equation \ref{eq:distribution_BRIEF}. For each sampling pair, the descriptor compares the intensity in the location of the first pixel "x" to that of second pixel "y" in patch "p"; if the intensity is greater then it assigns 1 in the final descriptor "\textit{f}" and 0 otherwise (Equation \ref{brief1})
\begin{equation}
\label{brief1}
f(p;x,y) = 
\begin{cases} 
      1 & p(x) < p(y) \\
      0 & otherwise
\end{cases}
\end{equation}
The comparisons result in a binary bitstring "$b_f$" (Equation \ref{brief2}).  
\begin{equation}
\label{brief2}
	b_f(p) = \sum_{1 \leq i \leq n_d} 2^{i-1} f(p;x,y) 
\end{equation}
The distance between two features can be computed very efficiently using the Hamming function.

\par
\textbf{Extensions of BRIEF:}
\begin{itemize}
\item ORB (Oriented fast and Rotated BRIEF) descriptor\cite{reference14} adds rotation invariance into existing BRIEF descriptor. Another contribution of ORB is the use of unsupervised learning in order to select optimal pixel location pairs.

\item BRISK (Binary Robust Invariant Scalable Keypoints)\cite{reference15} use concentric ring-based sampling patterns. BRISK compares two pixel pairs far from each other to compute the patch orientations and pixel pairs close to each other to compute the values of the descriptor itself.

\item FREAK (Fast REtinA Keypoint)\cite{reference16} also uses concentric rings arrangement with samples exponentially more pixel locations in the inner rings and uses unsupervised learning to choose optimal set of pixel location pairs.
\item LDB (Local Difference Binary)\cite{reference17} descriptor compares mean intensities in grids. LDB also compares the mean values of horizontal and vertical gradient differences.
\end{itemize}

LDA-Hash\cite{reference18} generates binary feature vectors yet the algorithm first computes SIFT and project the feature vector onto more discriminant space and uses a threshold in order to build binary vector. Thus, although the output is binary, the feature computation is substantially different from BRIEF and its variants, because LDA-Hash first extracts the SIFT descriptors and then generates binary vectors. Similarly, even though BinBoost \cite{trzcinski2013boosting} generates binary feature vectors by utilizing binary hash functions, it relies on computation of gradient orientations in rectangular image regions. 

\subsection{LATCH}
LATCH is the state-of-the-art binary descriptor giving comparable matching performance with best of non-binary descriptors \cite{reference1}. Though, BRIEF makes pre-smoothing of an image, this smoothing may cause to loss of high-frequency regions where key points are often detected. To overcome this issue, LATCH compares the patch triplets instead of pixel pairs. LATCH evaluates (equation \ref{latch}) the similarity of the "anchor" patch ($P_{t,a}$) to its two "companion" patches ($P_{t,1}$ and $P_{t,2}$) by computing their Frobenious norm and assigns 1 if Frobenious norm of difference of "anchor" and "companion1" patches is bigger than Frobenious norm of difference of "anchor" and "companion2" patches; otherwise assigns 0. LATCH takes the windows of size 48x48 around key points as BRIEF, and size of anchor and companion patches are fixed to 7x7.
\begin{equation}
\label{latch}
	g(W,s_t) = 
\begin{cases} 
      1 & \text{if } \lVert P_{t,a}-P_{t,1} \rVert_F^2 > \lVert P_{t,a}-P_{t,2} \rVert_F^2 \\
      0 & \text{otherwise}
\end{cases}
\end{equation}

LATCH does not sample the locations of patch triplets randomly. Instead, patch triplets are learned from Liberty, Notre Dame, and Yosemite dataset collections by supervised learning. C++ implementation of LATCH descriptor is available, and we based our experiments on this implementation.

\subsection{Other Color Descriptors}
Most of the extensions of existing feature extractors with color, "partially" use color attributes of image patches. \cite{reference4} handled color extensions of local descriptors for Video Concept Detection. They present a RGB-SURF, RGB-ORB, OpponentSURF and OpponentORB descriptors. In contrast to conventional descriptors, RGB-SURF and RGB-ORB apply original SURF or ORB descriptors directly to each of the RGB channels and concatenate feature vectors obtained from each channel, thus the size of the vector is tripled. 
In the case of OpponentSURF/OpponentORB, algorithm initially transform the RGB image to the opponent color space and use the transformed channels $O_1$, $O_2$ and $O_3$. $O_3$ is the luminance channel, i.e. the one that the original SURF/ORB descriptors use. The other two channels ($O_1$ and $O_2$) capture the color information, where $O_1$ is the red-green component and $O_2$ is the blue-yellow component. Similar with RGB-SURF/ORB, the original SURF or ORB descriptors are then applied separately to each transformed channel and the final feature vector is the concatenation of the three feature vectors extracted from the three channels independently.
\par
In \cite{reference5} a wide set of color descriptors and their invariant properties are examined for recognition accuracy. Color descriptors that are based on color histograms and SIFT based orientation histograms are compared. Koen van de Sande  et al. \cite{reference5}, state that SIFT descriptor is not invariant to light color changes, because the intensity channel, i.e. gray scale is a combination of the R, G and B channels. They proposed HSV-SIFT, where original SIFT descriptors is implemented to each of the channels of HSV color space. 

In general, most of the color approaches in feature extraction treat channels independently, missing the rich cross-channel statistics. Opponent color space alleviates this problem to some extent and show best results in object recognition, yet there is no mixing between red-green and blue-yellow components, more importantly the size of the vector needs to be tripled. Another reason for us to avoid opponent channel approach is for minimizing the computational load for extraction because it at least requires a color transformation operation to the whole image.

\section{Method} \label{section3}
\par
Conventional binary feature descriptors convert the image into gray scale, then binary comparisons are performed on a fixed size window around the detected keypoint. BRIEF descriptor uses randomly generated pixel location pairs and compare the grayscale value of these pairs. The image is smoothed before pixel value comparison for robustness to noise. Pixel value comparisons creates a binary feature vector, the size of which is determined by the total number of comparisons. Instead of single pixel value comparison at two locations, LATCH utilizes small image patches at three pixel locations. LATCH uses pixel patch triplets in order to avoid any potential loss of high frequencies in image.  LATCH compares the sum of pixel intensities of patch windows of size \big[7,7\big]. Both BRIEF and LATCH uses the detection windows of size \big[48,48\big] in their OpenCV implementations. We used the same parameters in our experiments for consistency. It should be noted that the both of the descriptors focus on intensity level of image in their comparisons. However RGB image carries both the color and intensity information. On the other hand YCbCr represents the intensity and color attributes better. We have focused on making comparisons, either BRIEF and LATCH, on Color Space in order to catch the color information in addition to intensity information. Another advantage of our approach is decreasing the computational cost by avoiding color transformation.

\subsection{Sampling in RGB Space} \label{subsection_3_1}
BRIEF and LATCH uses the grayscale pixel values in binary comparisons which loses the color information. However, a very straightforward extension of these descriptors can be designed which utilizes the color information. In order to use color property we made binary comparisons on the RGB image. We have used the same parameters with OpenCV implementations of BRIEF and LATCH. Additional advantage of this approach is not to transform the original image into gray scale, thus decreasing the computational cost. 
We call the RGB binary comparison as "3D RGB Sampling". 3D RGB Sampling is visualized in Fig.\ref{ColorSampling}. BRIEF uses the pixel locations that are sampled from an isotropic Gaussian distribution \eqref{eq:distribution_BRIEF}. Sampling from this distribution generates pixel locations that are closer to detected key points in 2D image.
\begin{equation}\label{eq:distribution_BRIEF}
\textbf{(X,Y)} \sim i.i.d.    Gaussian(0,\frac{1}{25}S^{2})
\end{equation}

\par
In addition to this distribution, we generate uniformly distributed numbers \{1,2,3\} that indicate the color channels. Instead of sampling pixel intensities in 2-Dimensional space \ref{ColorSampling}, we sample them in 3-Dimensional Space Fig.\ref{ColorSampling}, i.e. sampling is done among all channels of an image: R-channel, G-channel and B-channel. With this sampling approach, we create three dimensional pixel locations. 2D random distribution in Fig.2 is stretched among 3rd dimension and got the distribution Figured in Fig.\ref{ColorSampling}. In 3D Sampling, some comparisons might be made on the same color channel giving a similar comparison as in conventional binary descriptors in single images. On the contrary, the comparison might be cross-channel, e.g. the first could be chosen from R-channel, while the second one could be chosen from G-channel. Two points that are selected from the same channel is equivalent with 2D Sampling, while the points that are selected from different channels imply comparing the R-channel intensity with the intensity of the G-channel. Basically, cross-channel comparison asks the question "Whether the points are more reddish or more bluish or more greenish?". While 2D comparisons in BRIEF or latch descriptors answers the question "Which point's intensity is greater?". Existing binary descriptors are not seeking for any color-based information, while our approach extracts color information. With this simple modification we show dramatic improvement in feature matching performance. Some sample point selections and their meanings are given in Table \ref{RGB_sampling_meanings}. 
\par
In the case of color-enhanced LATCH descriptor, in a similar vein pixel patches are sampled from 3D RGB Space instead of grayscale patches.
\begin {table}[h!]
\caption{Various selections of pixel points used in 3D RGB space comparison \ref{ColorSampling}.}
\begin{center}
			\begin{tabular}{| p{1.4cm} | p{1.4cm} | p{9cm}|}
			\hline
			1\textsuperscript{th} Point & 2\textsuperscript{nd} Point & Explanation\\ \hline
			[5,5,1] & [8,8,1] & 1. Which pixel intensity is greater; [5,5] on R-channel or [8,8] on R-channel?\\ \hline
			[5,5,1] & [5,5,3] & 1. "Whether the pixel intensity at [5,5] on R-channel is greater than pixel intensity at [5,5] on B-channel or not?"\\
			& & 2. "Which point is more reddish/bluish?"  \\ \hline
			[3,3,2] & [17,22,3] & 1. "Whether the pixel intensity at [3,3] on R-channel is greater than pixel intensity at [17,22] on B-channel or not?" \\
			& & 2. "Which point is more greenish/bluish?" \\ \hline
			... & ... & ...\\ \hline
			\end{tabular}
			\end {center}
\label{RGB_sampling_meanings}
\end {table}

\subsection{Sampling in YCbCr Space} \label{subsection_3_2}
\begin{figure}[h!]
	\centering
		\includegraphics [width=10.24cm, height=3.3cm] {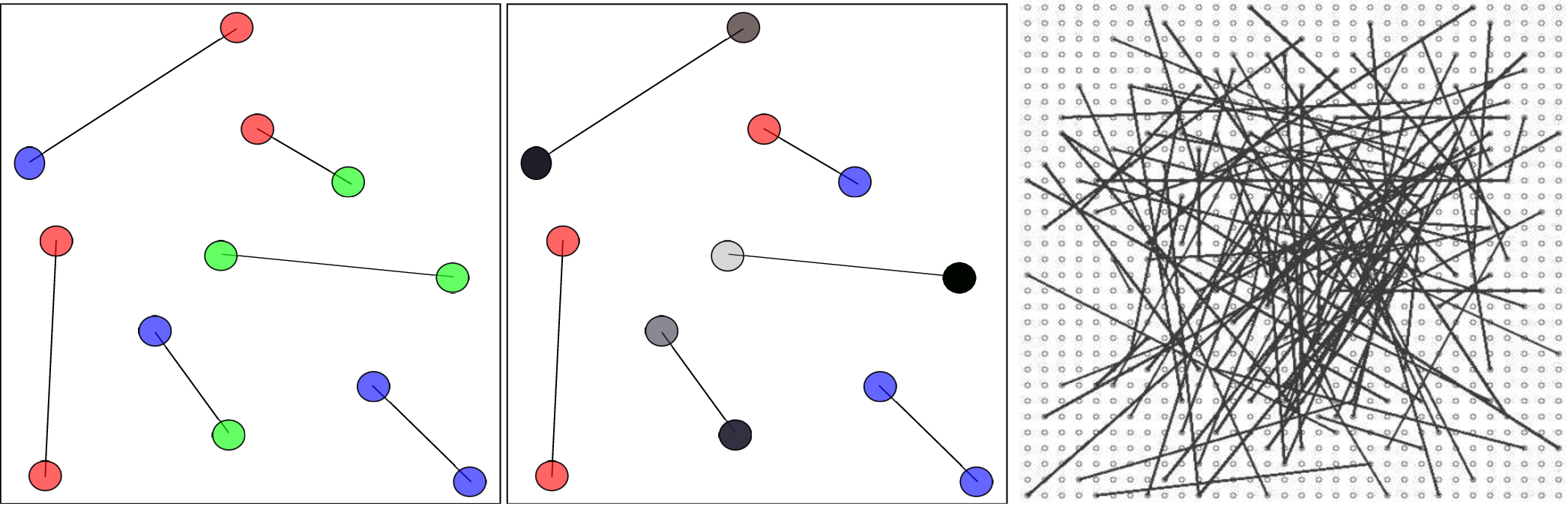}
\caption{\textbf{Left.} Points shown with red, green, blue colors indicate the pixel location that are taken from R-, G- and B-channels of RGB image, respectively. Stick with red and green colors at the ends, for example, compares the pixel intensity of R-channel with the pixel intensity G-channel of RGB image (see Table \ref{RGB_sampling_meanings}). \textbf{Middle.} Points shown with gray, blue and red points indicate the samples taken from Y-(luma), Cb-,Cr-(chroma) channels of YCbCr image, respectively. Tones of gray in Y-channel represent the pixel intensities in Y-channel. \textbf{Right.} 2D binary sampling pattern with the distribution shown in equation \ref{eq:distribution_BRIEF} used in \cite{reference2}.} 
\label {ColorSampling}
\end{figure}

RGB binary comparison ignores the intensity of pixels since it treats color values individually and does not take into account the grayscale value. While, 3D RGB Sampling improves the performance, it can be further improved by making binary comparisons in YCbCr color space. YCbCr represents  intensity and color attributes separately, so it will be very beneficial for using intensity and color at the same time. Conversion from RGB to YCbCr is almost equivalent to the cost of conversion from RGB to Gray. YCbCr separates the intensity (luma) from color information (chroma). "Y" component refers to luma component, while "Cb" and "Cr" refers to blue-difference and red-difference chroma components, respectively. In this color space we can handle the both luminance information and color information separately. That's why the performance of binary descriptors can be further improved by sampling in YCbCr color space as it is demonstrated in experimental results section\ref{experimental_results}. 3D YCbCr Sampling is visualized in Fig.\ref{ColorSampling}. 

\par
Luma component in YCbCr space represents the light intensity, while chroma represents the color. Luma component holds brightness information, therefore we sampled from Y-channel separately as in classical binary descriptors and sampled from Cb- and Cr-channels as we did in RGB 3D case explained above. We do not mix intensity and color comparisons. For color information extraction, channel numbers are randomly generated, that indicate the Cb and Cr channels.  So, in this YCbCr comparison, some number of comparisons are done in Y-channel only, and the other comparisons were done among Cb- and Cr-channels. 3D YCbCr Sampling is visualized in Fig.\ref{ColorSampling}. Examples of 3D YCbCr sampling and their explanations are given in Table \ref{YCbCr_sampling_meanings}. With this kind of comparison in color space, resultant bits of binary descriptors include color information in addition to intensity information. In 3D YCbCr Sampling, descriptor performance is improved further as it is demonstrated in experimental results section\ref{experimental_results}. Binary comparisons in YCbCr space gives the best results almost in all sequences. But, if computational cost is very crucial, then 3D RGB Sampling can also be used, while it performs significantly better than 2D Gray Sampling. 

\begin {table}[h]
\caption{Various selections of pixel points used in 3D YCbCr space comparison. Note that there is not any samplings among Y and Cb/Cr channels \ref{ColorSampling}}.
\begin{center}
			\begin{tabular}{| p{1.4cm} | p{1.4cm} | p{9cm}|}
			\hline
			1\textsuperscript{th} Point & 2\textsuperscript{nd} Point & Explanation\\ \hline
			[5,5,1] & [8,8,1] & 1. Which pixel's luminance value is greater; [5,5] on Y-channel or [8,8] on Y-channel?\\ \hline
						[3,3,2] & [17,22,2] & 1. "Whether the pixel chrominance at [3,3] on Cb-channel is greater than pixel chrominance at [17,22] on Cb-channel or not?" \\ \hline
			[5,5,2] & [5,5,3] & 1. "Whether the pixel chrominance at [5,5] on Cb-channel is greater than pixel chrominance at [5,5] on Cr-channel or not?"\\ 
			& & 2. "Which point is more bluish/reddish?"  \\ \hline

			... & ... & ...\\ \hline
			\end{tabular}
			\end {center}
\label{YCbCr_sampling_meanings}
\end {table}

\par
Very similar to the extension in RGB case, LATCH descriptor in YCbCr is done by sampling the pixel patches from either "Y" channel or among "Cb-", "Cr-" channels. Therefore, with 3D YCbCr sampling in LATCH descriptor we achieve simultaneous intensity/color comparisons avoiding the loss of high frequencies of the image by using patches instead of pixels.

\subsection {Feature Extraction Procedure \& Performance Evaluation} \label{subsection_3_3}
In experiments, feature extraction and matching procedures are taken from \cite{reference2}. Interest points were detected from the first image, and mapped into second image using the homography matrix, They are called interest points of the second image. Interest point detection in the second image is avoided in order not to account the repeatability performance of the detector in evaluating the descriptor performance. Step by step details of feature mapping and performance evaluations are given below:

\begin{enumerate}
\item Read two images that are going to be matched.
\item Detect the keypoints from first image using one of detectors and select \textbf{N} points among them. (detectors of SURF, ORB, CenCurE, SIFT, FAST, etc).
\item Map those points to second image using homography matrix. Refer these points as a interest points in the second image. (Not detecting keypoints from the second image)
\item Discard the points that are too close to borders of both first and second image. 
\item Extract the descriptor for keypoints locations of first image.
\item Extract the descriptor bits from the mapped pixel locations at the second image.
\item For a point in a first set of descriptors (image 1), search for the nearest descriptor in the second set of descriptors (image 2), and call them as a match.
\item Compute the number of correct matches, defined by the homography.
\item Evaluate the performance by dividing the \textbf{number of correct matches} by \textbf{N} -Total number of handled keypoints.
\end{enumerate}

Relative improvements due to color information are calculated as follows:

\begin{equation}\label{eq:performance}
	\textbf{RI}\% = \left(\dfrac{P' - P}{P}*100\%\right)
\end{equation}
"P - 2D Performance", "P' - 3D Performance" and "RI - Relative Improvement".

\par

\section{Experimental Results} \label{experimental_results}
We have used the Oxford dataset \footnote{\url{www.robots.ox.ac.uk/~vgg/research/affine}} in our experiments (whole set of results on the dataset are given as a supplementary document). Oxford dataset consist of 8 image sets with 5 different changes in imaging conditions:
\begin{itemize}
\item viewpoint changes (graf, wall imagesets)
\item scale changes (bark, boat imagesets)
\item image blur (bikes, trees imagesets)
\item JPEG compression (ubc imageset)
\item illumination (leuven imageset)
\end{itemize}
We used all imagesets except "boat" imageset because it is in grayscale, unable to give color information which is the purpose of this study. Detailed information about the sequences can be obtained from the Oxford dataset's site. Each set contains 6 images presenting increasing difficulty in the conditions stated above. We make descriptor matching between the first image and each of remaining five images. Feature decription and matching stages are stated in Section \ref{subsection_3_3}. We compared our approach with BRIEF and LATCH desciptors. In our experiments we did not exploit the "rotation invariance" attribute of LATCH, because we mapped the detected keypoint on the first image into second image by the use of homography matrix rather than detecting the keypoints from both images independently.  Gil \cite{reference1} showed that LATCH descriptor outperforms the BRIEF due to its robustness against "rotation" changes. But in our experiments, because we did not use "rotation invariance", LATCH sometimes falls behind BRIEF. Our purpose in showing results in LATCH without rotation invariance is presenting the generality of color inclusion into binary descriptors, whether it utilizes pixelwise or patchwise comparison. 

\par
\begin{figure}[h!]
	\centering
		\includegraphics [width=10.24cm, height=6.4cm] {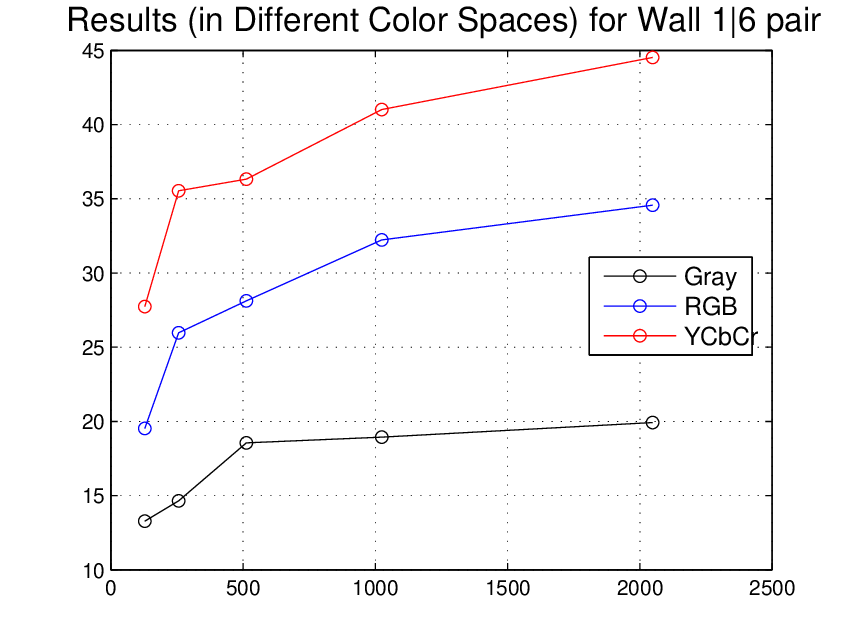}
\caption{Performances of Gray, RGB and YCbCr Samplings vs. Number of Binary Comparisons.}

\label {Bits}
\end{figure}
\par
\textbf{Performances as a function of the number of binary comparisons}\\
In \cite{reference2}, performance of BRIEF descriptor with respect to the number of binary comparisons (tests) is examined and it is observed that BRIEF requires at most  214 bits for Wall dataset to reach the performance of U-SURF descriptor, which requires 2048 bits. However, saturation can be observed in BRIEF for binary comparisons bigger than $\sim$512 bits. Our RGB-BRIEF and YCbCr-BRIEF outperforms the classical BRIEF at each number of tests (Figure \ref{Bits}). More importantly color enhanced binary descriptors do not saturate for small number of bits, but they go up. It leads to higher performances when we have computational power to handle more bits in binary feature extraction, thus it is looking to the future for faster hardware whereas classical binary descriptors will be unable to exploit them. Yet, if due to speed limitations it is required to use small number of bits, then our RGB-BRIEF and YCbCr-BRIEF still outperforms the classical BRIEF (Please see Supplementary Document for more results). 
\par
\begin{figure}[h!]
	\begin{center}
		\includegraphics [width=12.8cm, height=8cm] {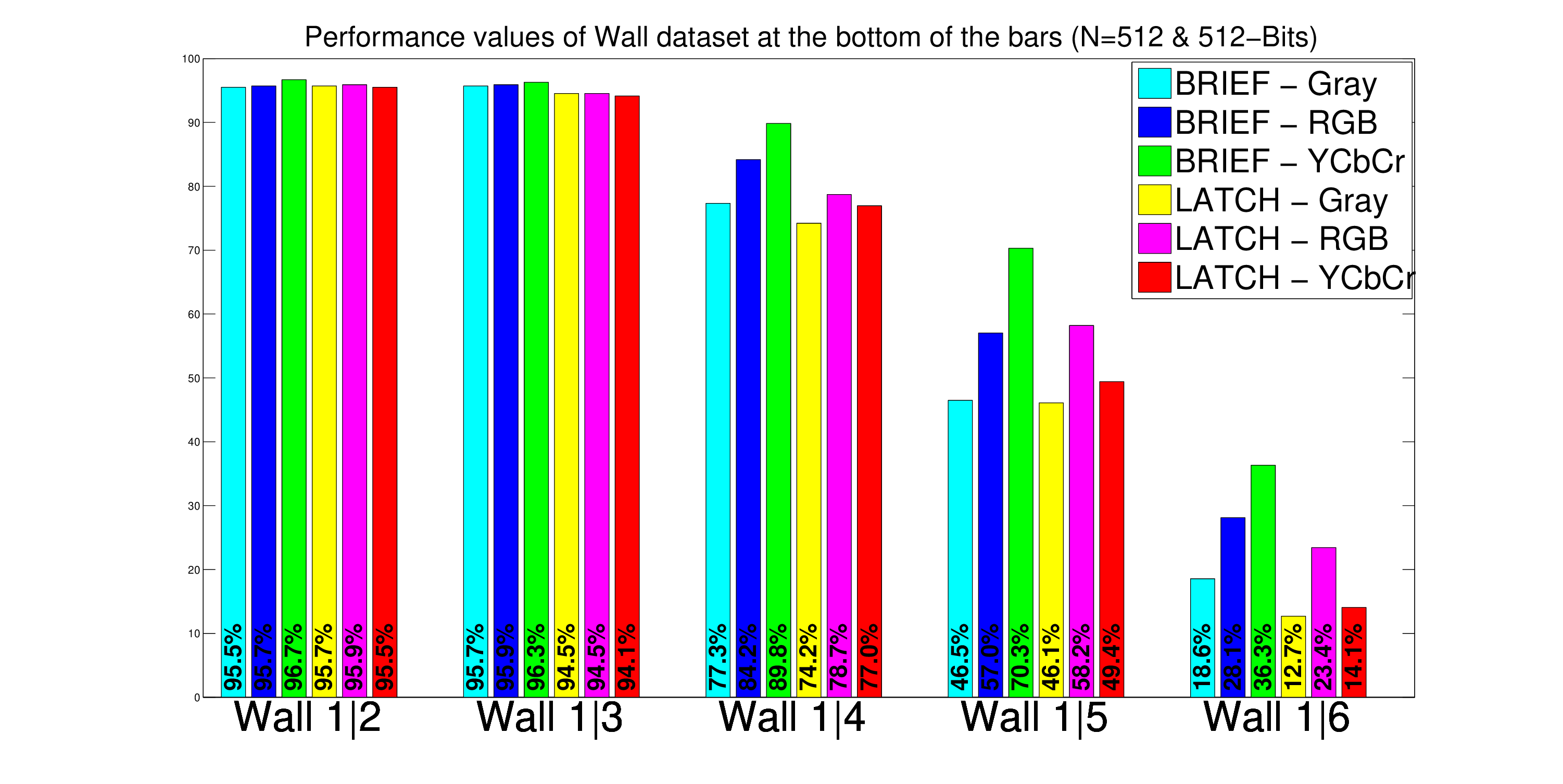}
		\includegraphics [width=12.8cm, height=8cm] {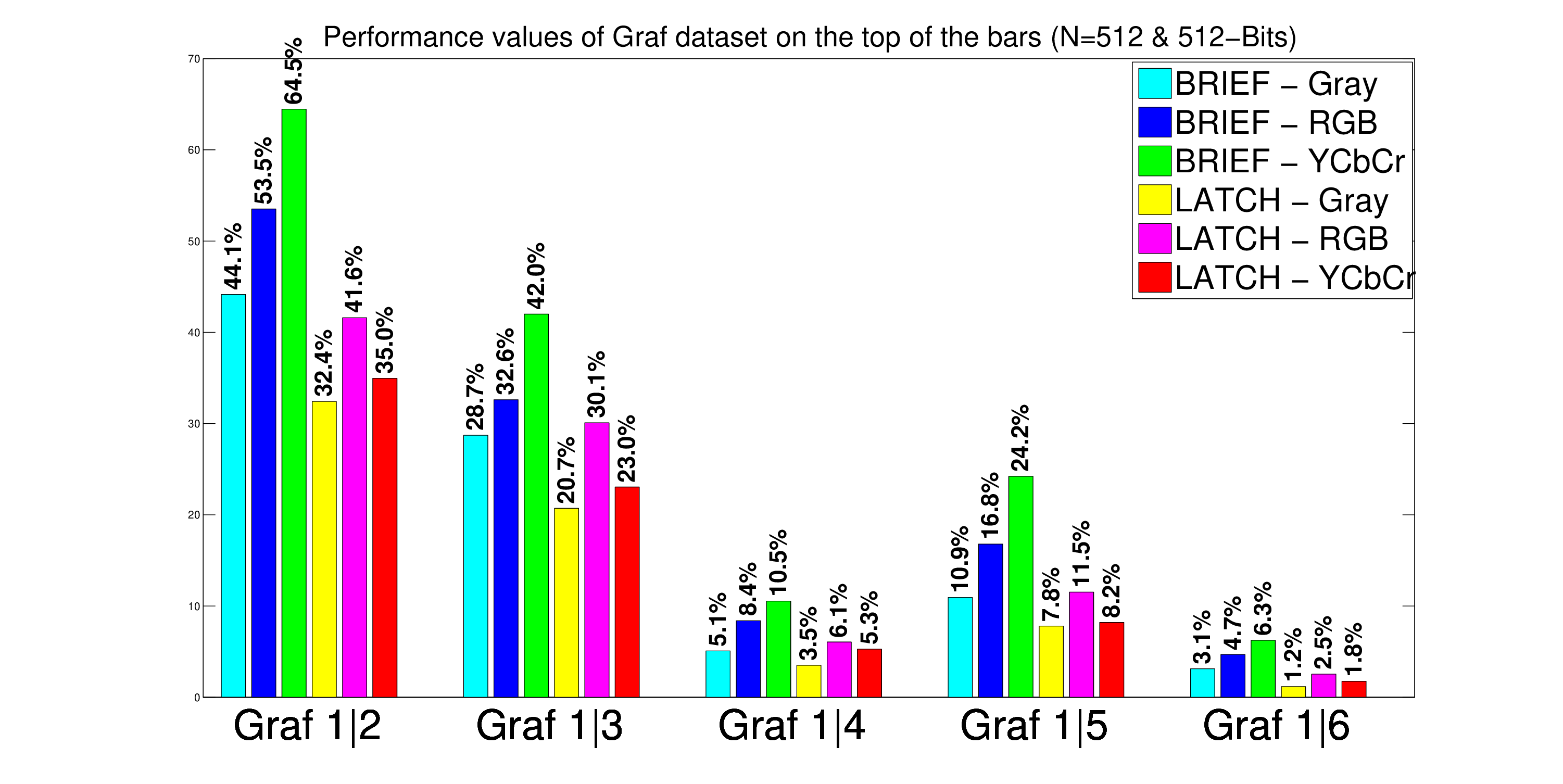}
	\end{center}
\caption{2D and 3D Binary Comparisons in Different Color Spaces: Wall and Graf imagesets. 512 Bits descriptors are used. N:Number of Keypoints to be matched.}
\label{wall}
\end{figure}

For detailed matching results, image matching performance between first and the rest of the images are given separately in bar charts (eg. 1\textbar2, 1\textbar3, 1\textbar4, 1\textbar5 and 1\textbar6 in Figure \ref{wall}) and Table \ref{Performances} for gray and and color descriptors. The matching performance drops with image number due to decreasing similarity with the first image. Please note that the term "Gray" for an algorithm corresponds to the original algorithm, i.e. BRIEF-Gray is the feature descriptor proposed in \cite{reference2}..

\begin{table}[h!]
\small
\caption{Performances of BRIEF and LATCH in different color spaces: \textbf{Gray(2D)}, \textbf{RGB (3D)} and \textbf{YCbCr (3D)} Sampling \\ 
\textbf{RGB-RI:} Relative Improvement in RGB space\ref{eq:performance} \\
\textbf{YCbCr-RI:} Relative Improvement in YCbCr space\ref{eq:performance}}

			\begin{tabular}{| p{1.45cm} | p{0.5cm} | p{0.5cm} | p{0.9cm} | p{0.45cm} | p{1.1cm} | p{0.5cm} | p{0.5cm} |p{0.9cm} | p{0.5cm} | p{1.3cm} |}
			\hline
			
			\textbf{\multirow{2}{*}{Dataset}} & \multicolumn{5}{c|}{\textbf{BRIEF (\%)}}  & \multicolumn{5}{c|}{\textbf{LATCH (\%)}} \\		\cline{2-11}
			\textbf{\multirow{2}{*}{}} & {\rotatebox[origin=c]{90}{\textbf{Gray}}} & \rotatebox[origin=c]{90}{\textbf{RGB}} & \multicolumn{1}{c|}{\rotatebox[origin=c]{90}{\textbf{RGB-RI }}}& \rotatebox[origin=c]{90}{\textbf{YCbCr}} & \rotatebox[origin=c]{90}{\textbf{YCbCr-RI }} & {\rotatebox[origin=c]{90}{\textbf{Gray}}} & {\rotatebox[origin=c]{90}{\textbf{RGB}}} & {\rotatebox[origin=c]{90}{\textbf{RGB-RI}}} & \rotatebox[origin=c]{90}{\textbf{YCbCr}} & \rotatebox[origin=c]{90}{\textbf{YCbCr-RI }}\\ \hline \hline
			\textbf{wall 1\textbar2} & 95.5 & 95.7 & \textbf{$\sim$0\%} & 96.7 & \textbf{$\sim$1\%} & 95.7 & 95.9 & \textbf{$\sim$0\%} & 95.5 & \textbf{$\sim$0\%} \\ \hline
			
			\textbf{wall 1\textbar3} & 95.7 & 95.9 & \textbf{$\sim$0\%} & 96.3 & \textbf{$\sim$1\%} & 94.5 & 94.5 & \textbf{$\sim$0\%} & 94.1 & \textbf{$\sim$0\%} \\ \hline
			
			\textbf{wall 1\textbar4} & 77.3 & 84.2 & \textbf{$\sim$9\%} & 89.8 & \textbf{$\sim$16\%} & 74.2 & 78.7 & \textbf{$\sim$6\%} & 77.0 & \textbf{$\sim$4\%} \\ \hline
			
			\textbf{wall 1\textbar5} & 46.5 & 57.0 & \textbf{$\sim$23\%} & 70.3 & \textbf{$\sim$51\%} & 46.1 & 58.2 & \textbf{$\sim$26\%} & 49.4 & \textbf{$\sim$7\%} \\ \hline
			
			\textbf{wall 1\textbar6} & 18.6 & 28.1 & \textbf{$\sim$51\%} & 36.3 & \textbf{$\sim$95\%} & 12.7 & 23.4 & \textbf{$\sim$84\%} & 14.1 & \textbf{$\sim$11\%} \\ \hline \hline
			\textbf{graf 1\textbar2} & 44.1 & 53.5 & \textbf{$\sim$21\%} & 64.5 & \textbf{$\sim$46\%} & 32.4 & 41.6 & \textbf{$\sim$28\%} & 35.0 & \textbf{$\sim$8\%} \\ \hline
			
			\textbf{graf 1\textbar3} & 28.7 & 32.6 & \textbf{$\sim$14\%} & 42.0 & \textbf{$\sim$46\%} & 20.7 & 30.1 & \textbf{$\sim$45\%} & 23.0 & \textbf{$\sim$11\%} \\ \hline
			
			\textbf{graf 1\textbar4} & 5.1 & 8.4 & \textbf{$\sim$65\%} & 10.5 & \textbf{$\sim$106\%} & 3.5 & 6.1 & \textbf{$\sim$74\%} & 5.3 & \textbf{$\sim$51\%} \\ \hline
			
			\textbf{graf 1\textbar5} & 10.9 & 16.8 & \textbf{$\sim$54\%} & 24.2 & \textbf{$\sim$122\%} & 7.8 & 11.5 & \textbf{$\sim$47\%} & 8.2 & \textbf{$\sim$5\%} \\ \hline
			
			\textbf{graf 1\textbar6} & 3.1 & 4.7 & \textbf{$\sim$52\%} & 6.3 & \textbf{$\sim$103\%} & 1.2 & 2.5 & \textbf{$\sim$108\%} & 1.8 & \textbf{$\sim$50\%} \\ \hline \hline
			\textbf{trees 1\textbar2} & 78.1 & 80.5 & \textbf{$\sim$3\%} & 85.2 & \textbf{$\sim$9\%} & 87.9 & 91.2 & \textbf{$\sim$4\%} & 88.9 & \textbf{$\sim$1\%} \\ \hline
			
			\textbf{trees 1\textbar3} & 66.0 & 72.7 & \textbf{$\sim$10\%} & 78.3 & \textbf{$\sim$19\%} & 84.8 & 86.5 & \textbf{$\sim$2\%} & 85.5 & \textbf{$\sim$1\%} \\ \hline
			
			\textbf{trees 1\textbar4} & 47.1 & 53.3 & \textbf{$\sim$13\%} & 57.8 & \textbf{$\sim$23\%} & 70.3 & 73.2 & \textbf{$\sim$4\%} & 70.9 & \textbf{$\sim$1\%} \\ \hline
			
			\textbf{trees 1\textbar5} & 62.1 & 65.2 & \textbf{$\sim$5\%} & 66.6 & \textbf{$\sim$7\%} & 74.4 & 76.0 & \textbf{$\sim$2\%} & 75.4 & \textbf{$\sim$1\%} \\ \hline
			
			\textbf{trees 1\textbar6} & 50.8 & 57.4 & \textbf{$\sim$12\%} & 66.2 & \textbf{$\sim$30\%} & 72.3 & 73.8 & \textbf{$\sim$2\%} & 72.3 & \textbf{$\sim$0\%} \\ \hline \hline
			\textbf{ubc 1\textbar2} & 100 & 100 & \textbf{$\sim$0\%} & 99.8 & \textbf{$\sim$0\%} & 100 & 100 & \textbf{$\sim$0\%} & 99.8 & \textbf{$\sim$0\%} \\ \hline
			\textbf{ubc 1\textbar3} & 100 & 100 & \textbf{$\sim$0\%} & 100 & \textbf{$\sim$0\%} & 100 & 100 & \textbf{$\sim$0\%} & 98.4 & \textbf{$\sim$0\%}  \\ \hline
			\textbf{ubc 1\textbar4} & 100 & 99.8 & \textbf{$\sim$0\%} &  99.4 & \textbf{$\sim$0\%} & 100 & 100 & \textbf{$\sim$0\%} & 96.7 & \textbf{$\sim$-3\%} \\ \hline
			\textbf{ubc 1\textbar5} & 97.9 & 97.3 & \textbf{$\sim$0\%} & 95.1 & \textbf{$\sim$-3\%} & 99.6 & 99.4 & \textbf{$\sim$0\%} & 95.5 &  \textbf{$\sim$-4\%} \\ \hline
			\textbf{ubc 1\textbar6} & 94.7 & 95.5 & \textbf{$\sim$1\%} & 93.2 & \textbf{$\sim$-2\%} & 97.9 & 97.9 & \textbf{$\sim$0\%} & 93.2 & \textbf{$\sim$-5\%} \\ \hline \hline
			\textbf{leuven1\textbar2} & 95.7 & 95.5 & \textbf{$\sim$0\%} & 95.5 & \textbf{$\sim$0\%} & 97.3 & 96.5 & \textbf{$\sim$-1\%} & 97.5 & \textbf{$\sim$0\%} \\ \hline
			
			\textbf{leuven1\textbar3} & 94.7 & 94.3 & \textbf{$\sim$0\%} & 95.5 & \textbf{$\sim$1\%} & 95.9 & 94.3 & \textbf{$\sim$-2\%} & 95.3 & \textbf{$\sim$-1\%} \\ \hline
			
			\textbf{leuven1\textbar4} & 94.1 & 91.8 & \textbf{$\sim$-2\%} & 94.5 & \textbf{$\sim$0\%} & 95.1 & 93.6 & \textbf{$\sim$-2\%} & 94.9 & \textbf{$\sim$0\%} \\ \hline
			
			\textbf{leuven1\textbar5} & 92.2 & 90.2 & \textbf{$\sim$-2\%} & 91.8 & \textbf{$\sim$0\%} & 95.1 & 91.6 & \textbf{$\sim$-4\%} & 93.4 & \textbf{$\sim$-2\%} \\ \hline
			
			\textbf{leuven1\textbar6} & 89.5 & 88.3 & \textbf{$\sim$-1\%} & 89.6 & \textbf{$\sim$0\%} & 93.2 & 90.6 & \textbf{$\sim$-3\%} & 91.6 & \textbf{$\sim$-2\%} \\ \hline \hline
			\textbf{bikes 1\textbar2} & 94.5 & 94.5 & \textbf{$\sim$0\%} & 94.7 & \textbf{$\sim$0\%} & 96.3 & 96.1 & \textbf{$\sim$0\%} & 97.1 & \textbf{$\sim$1\%} \\ \hline
			\textbf{bikes 1\textbar3} & 95.7 & 94.7 & \textbf{$\sim$-1\%} & 94.7 & \textbf{$\sim$-1\%} & 96.1 & 96.1 & \textbf{$\sim$0\%} & 96.3 & \textbf{$\sim$0\%} \\ \hline
			\textbf{bikes 1\textbar4} & 95.3 & 95.1 & \textbf{$\sim$0\%} & 93.2 & \textbf{$\sim$-2\%} & 95.5 & 96.1 & \textbf{$\sim$1\%} & 95.9 & \textbf{$\sim$0\%} \\ \hline
			\textbf{bikes 1\textbar5} & 92.0 & 94.9 & \textbf{$\sim$3\%} & 91.6 & \textbf{$\sim$0\%} & 95.7 & 95.1 & \textbf{$\sim$-1\%} & 96.5 & \textbf{$\sim$1\%} \\ \hline
			\textbf{bikes 1\textbar6} & 86.3 & 86.3 & \textbf{$\sim$0\%} & 85.0 & \textbf{$\sim$-2\%} & 90.0 & 88.7 & \textbf{$\sim$-1\%} & 91.2 & \textbf{$\sim$1\%} \\ \hline \hline
			\textbf{bark 1\textbar2} & 4.3 & 11.7 & \textbf{$\sim$172\%} & 18.0 & \textbf{$\sim$318\%} & 2.0 & 7.0 & \textbf{$\sim$250\%} & 3.5 & \textbf{$\sim$75\%} \\ \hline
			\textbf{bark 1\textbar3} & 0.0 & 0.4 & \textbf{-} & 0.2 & \textbf{-} & 0.2 & 0.2 & \textbf{$\sim$0\%} & 0.4 & \textbf{$\sim$100\%} \\ \hline
			\textbf{bark 1\textbar4} & 0.0 & 0.2 & \textbf{-} & 0.0 & \textbf{-} & 0.4 & 0.0 & \textbf{-} & 0.4 & \textbf{-} \\ \hline
			\textbf{bark 1\textbar5} & 0.1 & 0.8 & \textbf{$\sim$700\%} & 1.6 & \textbf{$\sim$1500\%} & 0.2 & 1.0 & \textbf{$\sim$400\%} & 0.2 & \textbf{$\sim$-80\%} \\ \hline
			\textbf{bark 1\textbar6} & 0.2 & 0.4 & \textbf{$\sim$100\%} & 0.8 & \textbf{$\sim$300\%} & 0.2 & 0.4 & \textbf{$\sim$100\%} & 0.2 & \textbf{$\sim$-100\%} \\ \hline

			\end{tabular}
\label{Performances}
\end {table}

\textbf{1. Wall imageset:} Comparative results are shown in Figure \ref{wall}.
	\begin{itemize}
		\item wall1\textbar2: Both BRIEF and LATCH were very successful. Color-enhanced features gives the same, even better results. Matching score are written at the bottom of each bar. \footnote{Due to ignorance of rotation invariance of LATCH, its matching results are worse compared to matching scores of BRIEF both in Gray and Color cases.}

	\item wall1\textbar3: Similar results with wall1\textbar2.
	\item wall1\textbar4: Since viewpoint changes more, color-enhanced BRIEF and LATCH gives significantly better performances. Gray BRIEF gives 77\% performance, while RGB BRIEF and YCbCr BRIEF gives the 84\% and 90\%, respectively. This improvement is due to the usage of color which helps to withstand against changes in viewpoint. 
	\item wall1\textbar5: The improvement due to color is much more visible in this case. The relative performance improvement is 12\% and 51\% for RGB-BRIEF and YCbCr-BRIEF respectively.  

\item wall1\textbar6: Improvements become dramatic because of much challenging viewpoint change. RGB-BRIEF shows 51\% and YCbCr-BRIEF shows  95\% matching improvement.
\end{itemize}

\par
\textbf{2. Graf imageset: Figure \ref{wall}}
\par
In Graf imageset, impact of color information can be observed much more clearly. The improvement in matching performance is experienced even in matching the first and second images. RGB-BRIEF relative improvements are: 21\%, 14\%, 65\%, 54\%, 52\% for each successive matchings. YCbCr-BRIEF's relative improvements are: 46\%, 46\%, 102\%, 122\%, and 103\% respectively. The performance is \textbf{doubled} by the usage of color information, which does not add any computational cost.

\begin{figure}[h!]
	\begin{center}
		\includegraphics [width=12.8cm, height=8cm] {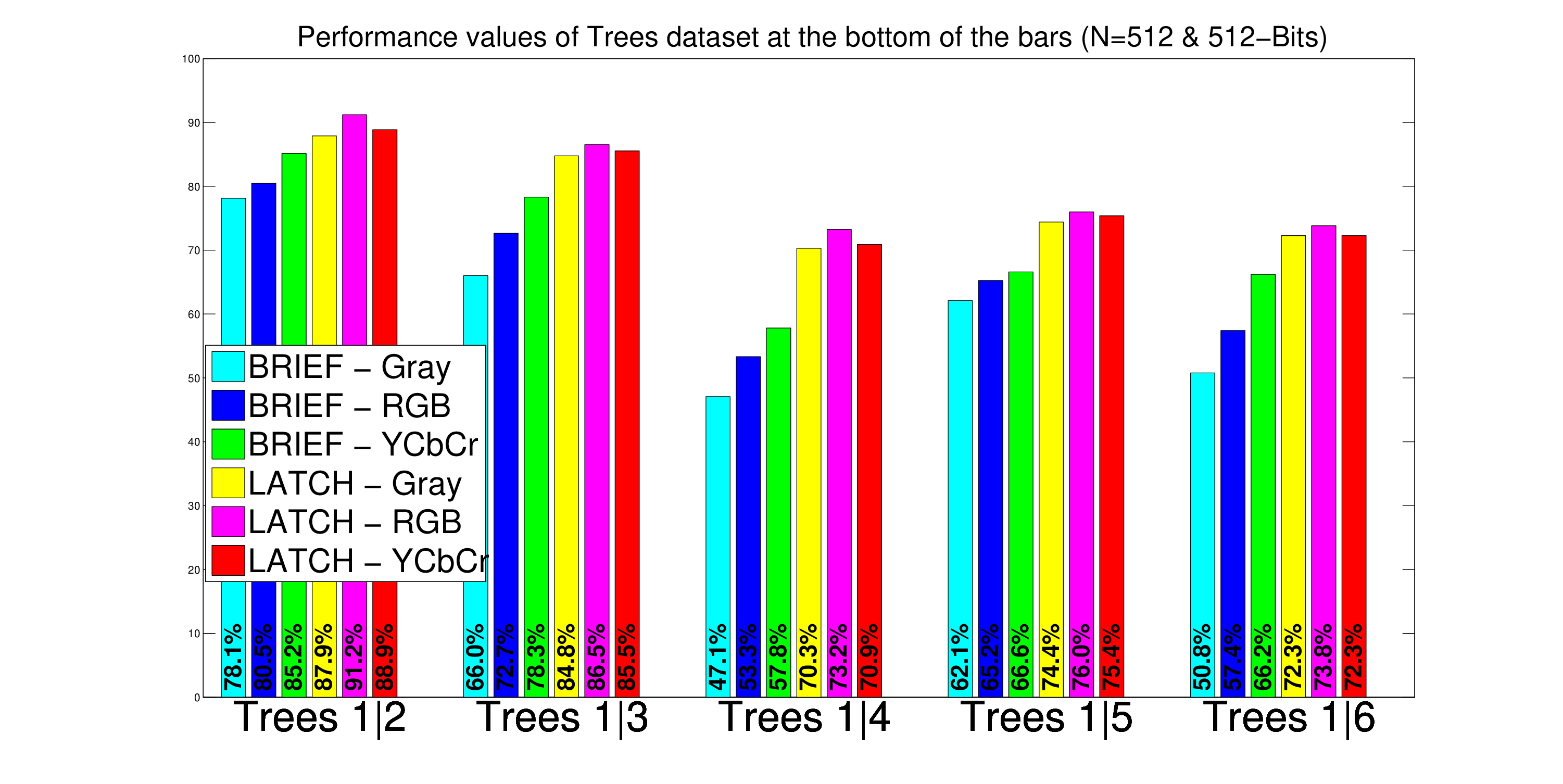}
	\end{center}
\caption{2D and 3D Binary Comparisons in Different Color Spaces: trees imageset. 512 Bits descriptors are used. N:Number of Keypoints to be matched.}
\label {trees}
\end{figure}

\par
\textbf{3. Trees imageset: Figure \ref{trees}}
\par
Again we observe significant improvements in performance. RGB-BRIEF relative improvements are: 21\%, 14\%, 65\%, 54\%, 52\% for each successive matchings. YCbCr-BRIEF's relative improvements are: 46\%, 46\%, 102\%, 122\%, and 103\% respectively. \footnote{Since imagining condition in "trees" imageset is blur rather than rotation (Note that we turned off the rotation invariance of LATCH descriptor), LATCH outperforms the BRIEF in this imageset, while in previous two imagesets it does not. }

\par
\textbf{4. Ubc, Bark, Leuven and Bike imagesets: \footnote{Figures are given in Supplementary Document}}
\par
In three of the rest of the imagesets (Ubc, Leuven, Bike), BRIEF and LATCH descriptor performance saturate at very high matching scores. It is crucial that color descriptors do not fall behind the gray scale versions in the case of saturation. The performance for Bark set is very low even for the first two images, yet color information adds significant improvement (please see Supplementary Document for the bar plots).

\clearpage
\section{Discussion}
We offer a very simple extension of existing binary descriptors: make binary comparison in 3D color image volume, not the 2D grayscale image. It is interesting that this idea was not tested before. We show dramatic matching improvements in standart imagesets. If comparison is performed in RGB space, it also yields a speedup due to the fact that grayscale conversion can be avoided. If YCbCr color space is used, performance is even better, and the computational cost is almost identical to 2D-Gray binary descriptors. 

YCbCr color space sampling is superficially similar to OpponentORB \cite{reference4,reference5}. However, even though opponent channels mix color information to some extent, each channel is still treated separately. Yet, random binary comparisons in Cb/Cr fully mix the color information due to intrachannel and interchannel components. 

In the future, we are planning to learn optimal sampling patterns \cite{trzcinski2012efficient,reference14,trzcinski2013boosting} in color spaces. This optimization might be more crucial in color than grayscale because of the curse of dimensionality. We would like to make more comprehensive tests on the color binary descriptors, specifically on object/scene detection applications.   

\section{Acknowledgments}
This research is supported by The Scientific and Technological Research Council of
Turkey (TUB\.{I}TAK) Career Grant, No: 114E554.

\end{document}


\pagestyle{fancy}
\rhead{SUPPLEMENTARY DOCUMENT}
\lhead{U-CATCH}

\title{SUPPLEMENTARY DOCUMENT: \newline U-CATCH: Using Color ATtribute of image patCHes in Binary Descriptors}

\author[1]{Ozgur Yilmaz}
\author[2]{Alisher Abdulkhaev}
\affil[1,2]{Turgut \"{O}zal University, Department of Computer Engineering, Ankara Turkey}

\maketitle

\begin{center}
\noindent\rule{10cm}{0.5pt}
\end{center}

\section{Performances of other image sets (Ubc, Bark, Leuven, Bikes) in different color spaces.}
\par
\begin{figure}[]
	\begin{center}
		\includegraphics [width=12.8cm, height=7.5cm] {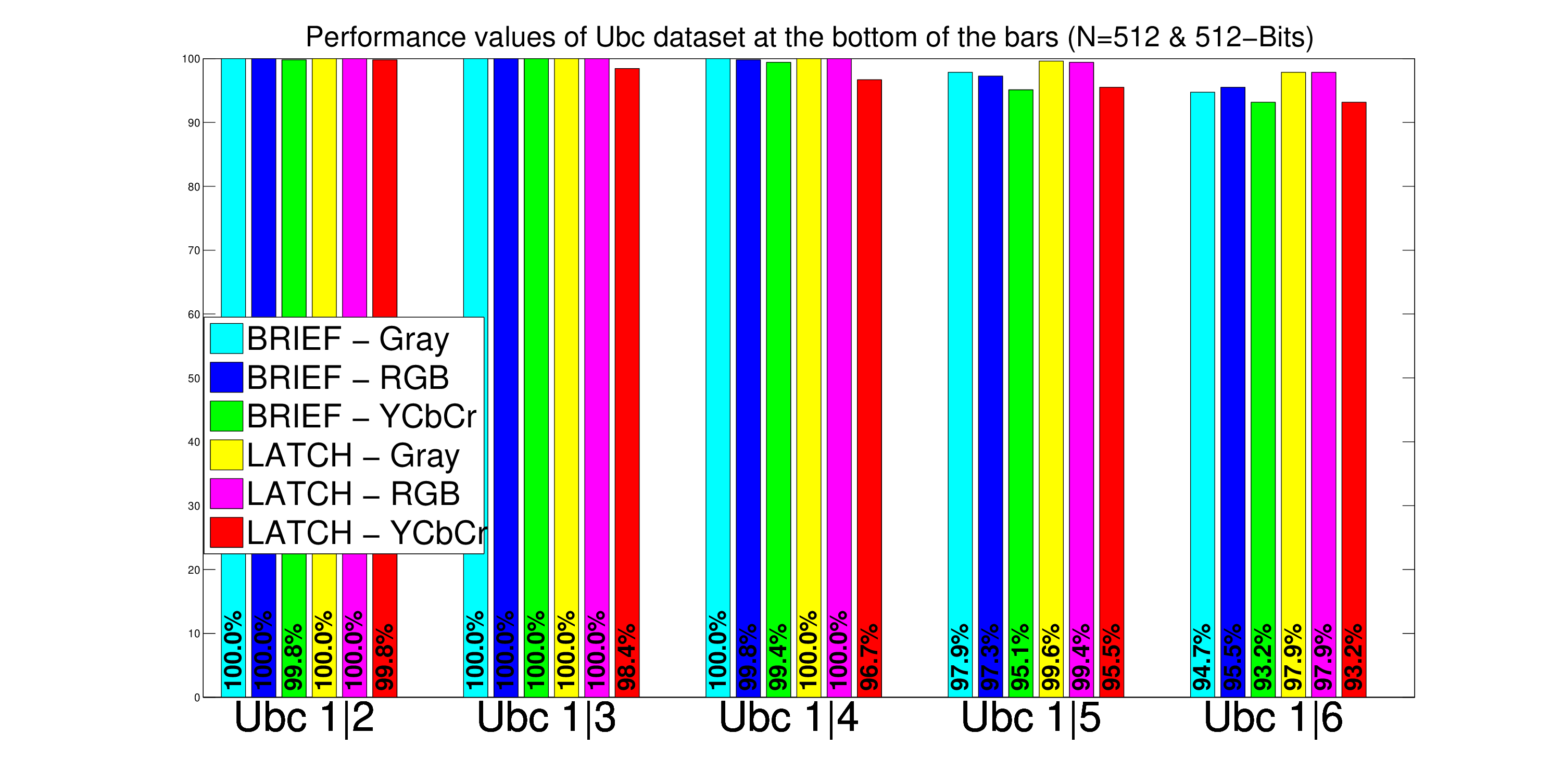}
		\includegraphics [width=12.8cm, height=7.5cm] {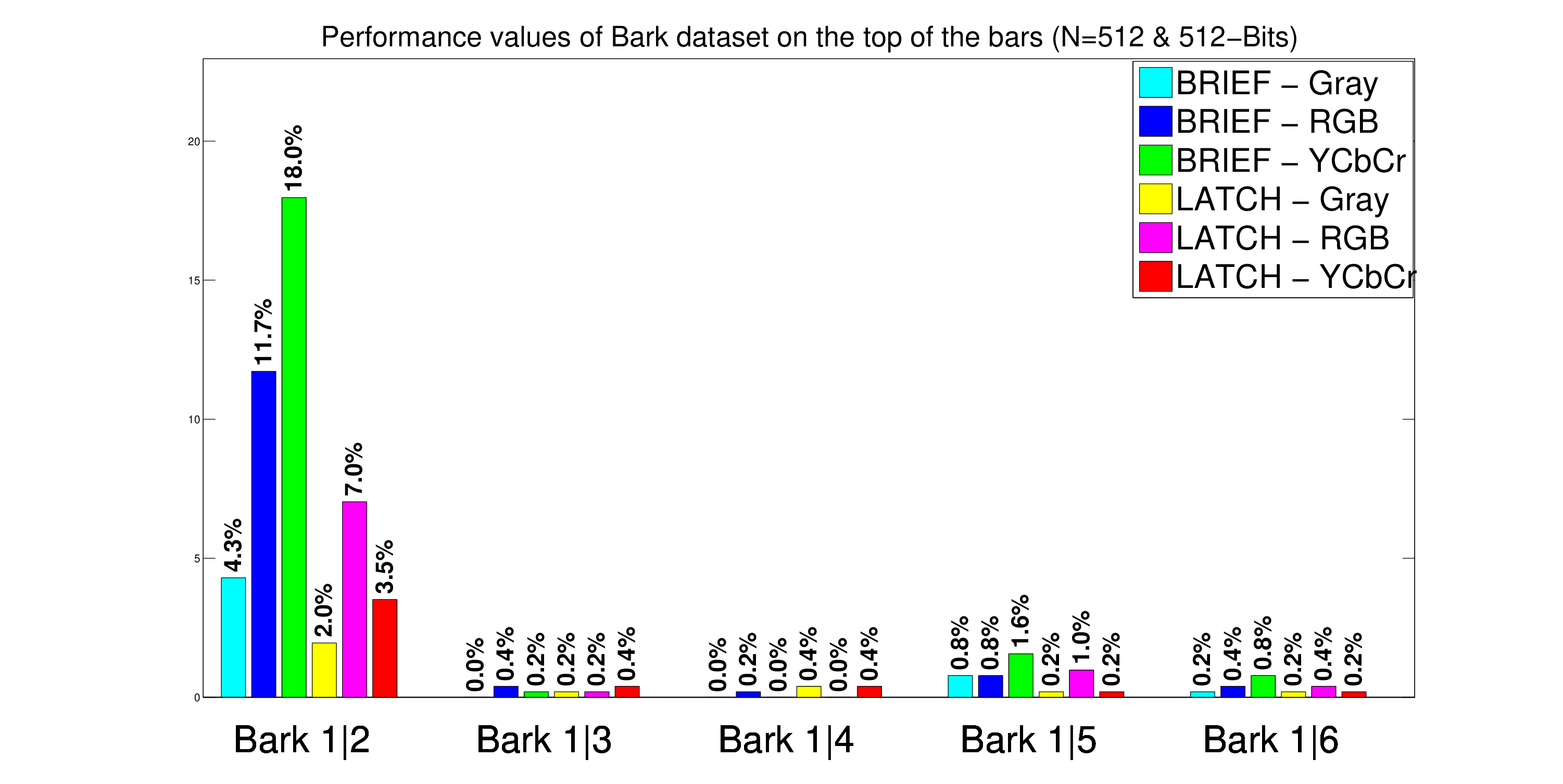}
	\end{center}
\caption{2D and 3D Binary Comparisons in Different Color Spaces: Ubc and Bark imagesets. 512 Bits descriptors are used. N:Number of Keypoints to be matched.}
\label{ubc}
\end{figure}

Performance of gray descriptors, both BRIEF and LATCH, already reaches to a very high level on each Ubc\ref{ubc} image pairs. RGB and YCbCr descriptors also perform well on image pairs of Ubc image set. Differences between 2D and 3D samplings seem to be insignificant.
\par 
In Bark image set\ref{ubc}, 3D-Color sampling (RGB-BRIEF, YCbCr-BRIEF, RGB-LATCH, YCbCr-LATCH) outperform 2D-Gray versions, significantly. In Bark1\textbar3 and Bark1\textbar4 pairs Gray-BRIEF could not catch any true matches, while RGB-BRIEF and YCbCr-BRIEF gave 0.4\% and 0.2\% of true matches, respectively. 
\par
\begin{figure}[]
	\begin{center}
		\includegraphics [width=12.8cm, height=7.5cm] {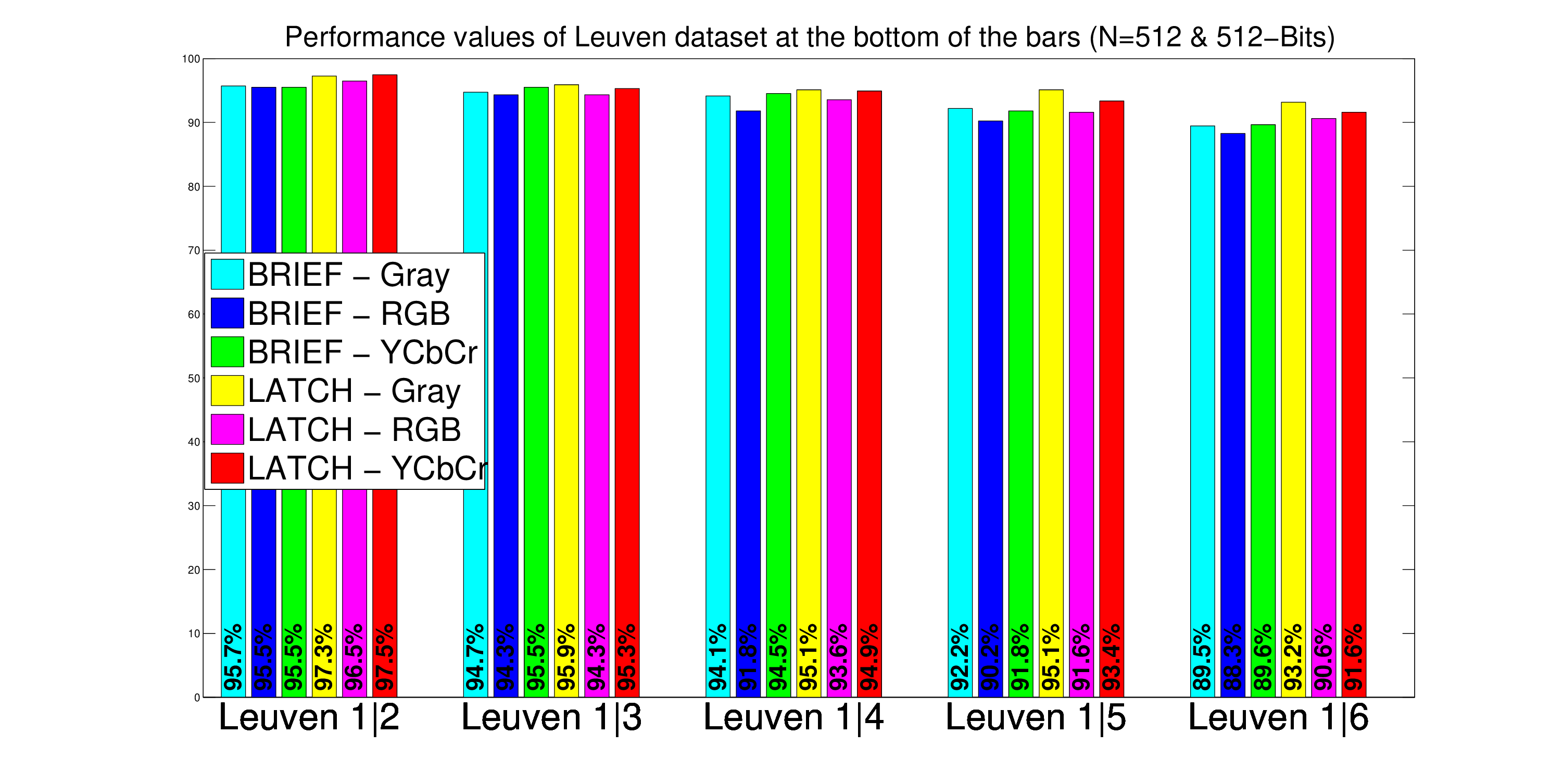}
		\includegraphics [width=12.8cm, height=7.5cm] {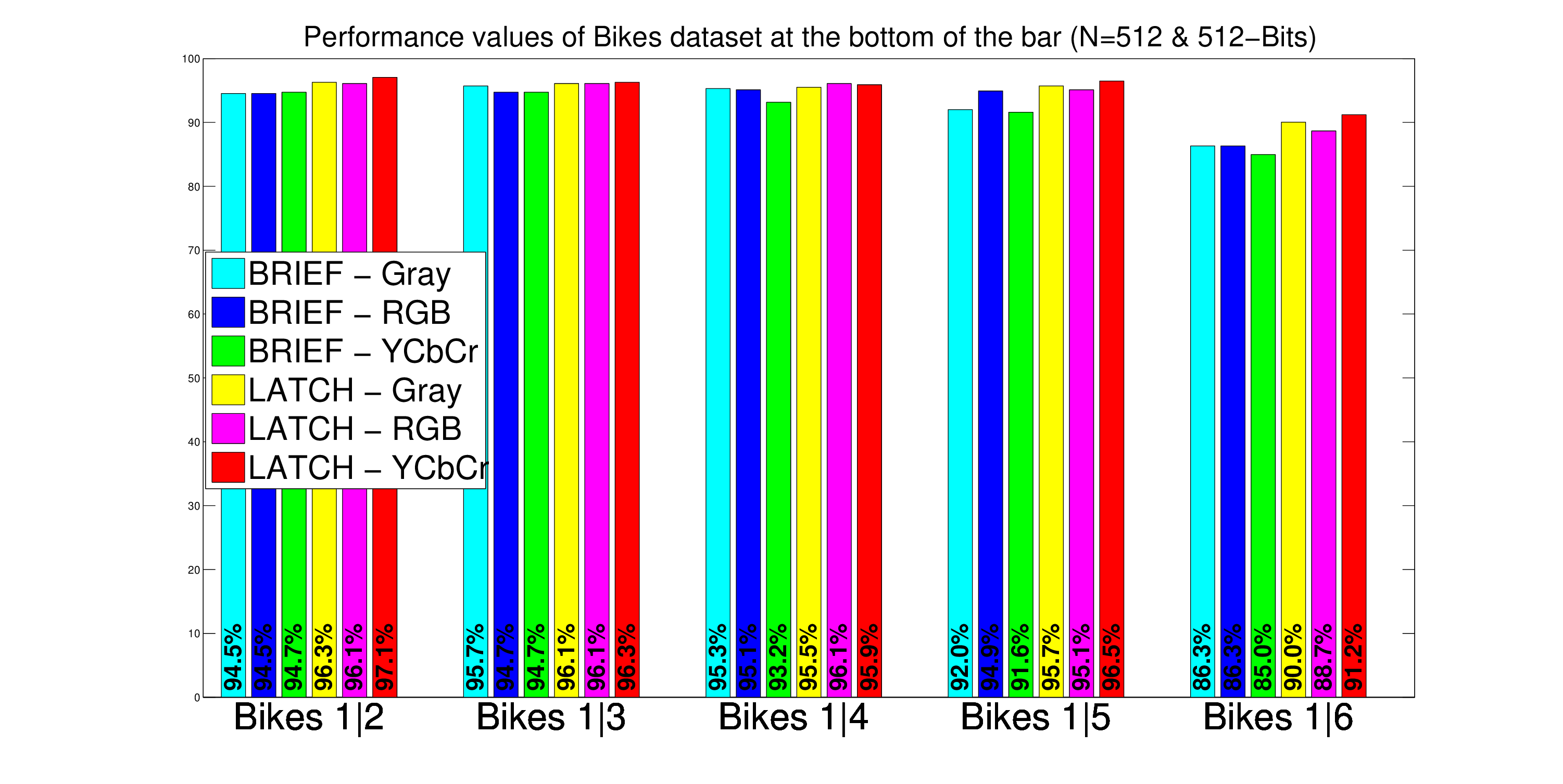}
	\end{center}
\caption{2D and 3D Binary Comparisons in Different Color Spaces: Leuven and Bikes imageset. 512 Bits descriptors are used. N:Number of Keypoints to be matched.}
\label{leuven}
\end{figure}

In the rest two image sets, Leuven and Bikes, all types of descriptors give the similar performances as shown in \ref{leuven}.

\section{Performances of Gray, RGB and YCbCr descriptors with respect to number of tests.}
\begin{figure}[]
	\begin{center}
		\includegraphics [width=12.8cm, height=8cm] {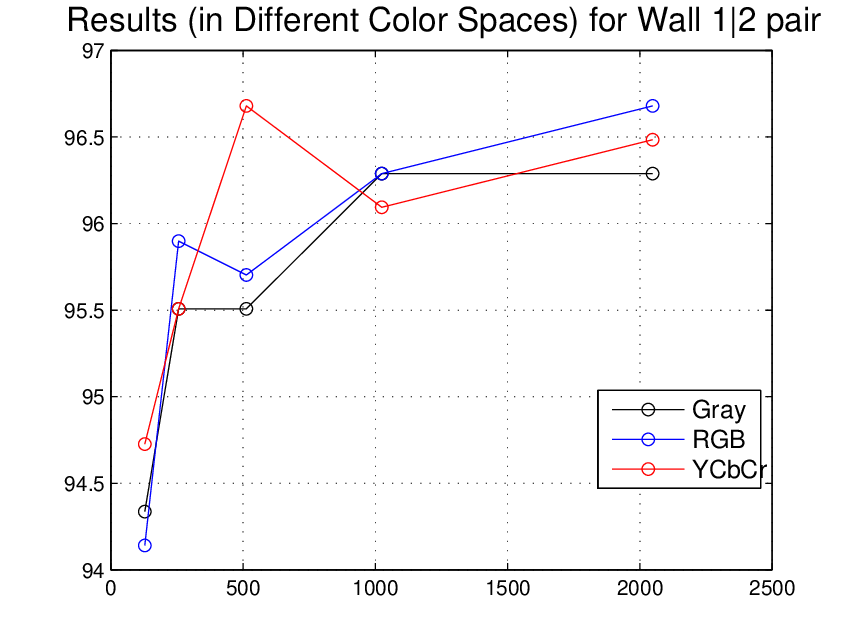}
		\includegraphics [width=12.8cm, height=8cm] {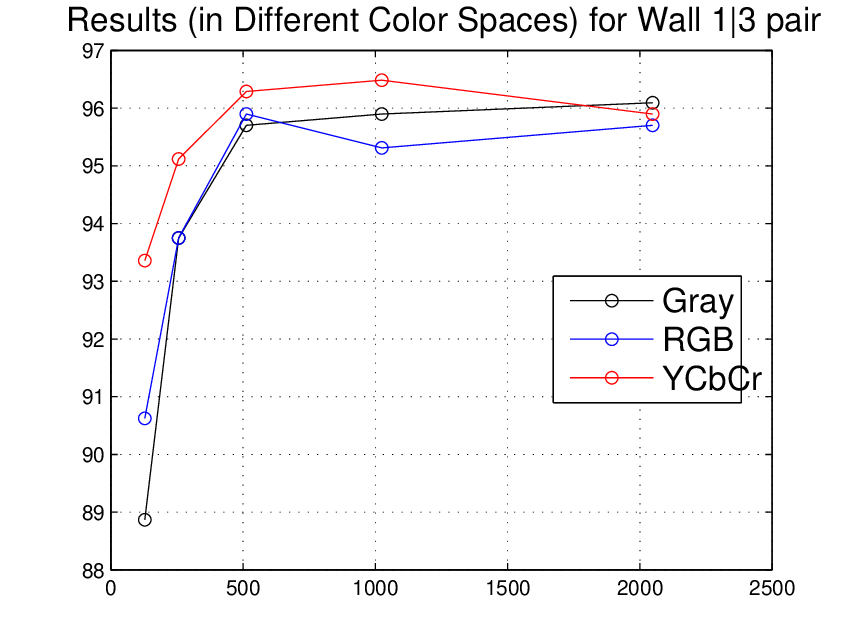}
	\end{center}
\caption{Performances of Gray, RGB and YCbCr Samplings vs. Number of Samples.}
\label{wall12_wall13}
\end{figure}

\begin{figure}[h!]
	\begin{center}
		\includegraphics [width=12.8cm, height=8cm] {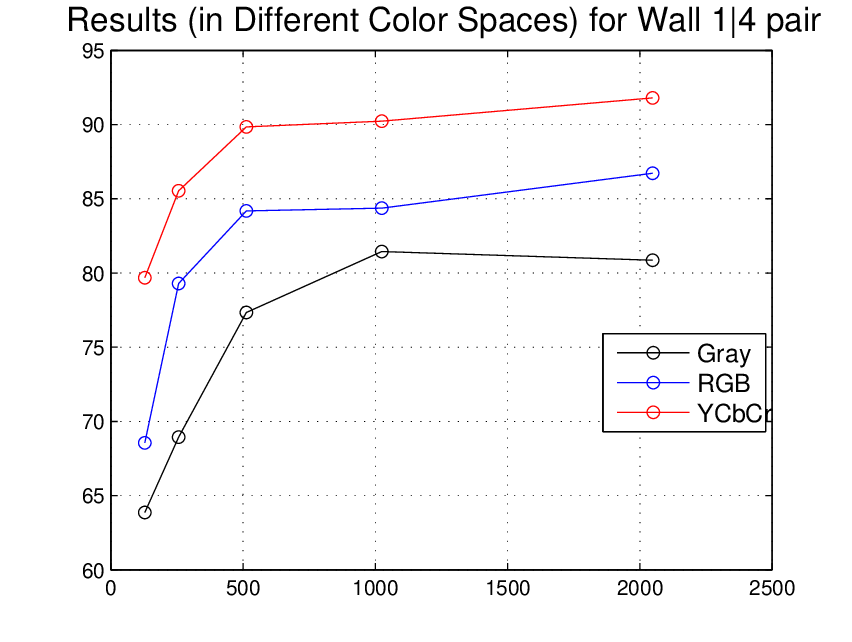}
		\includegraphics [width=12.8cm, height=8cm] {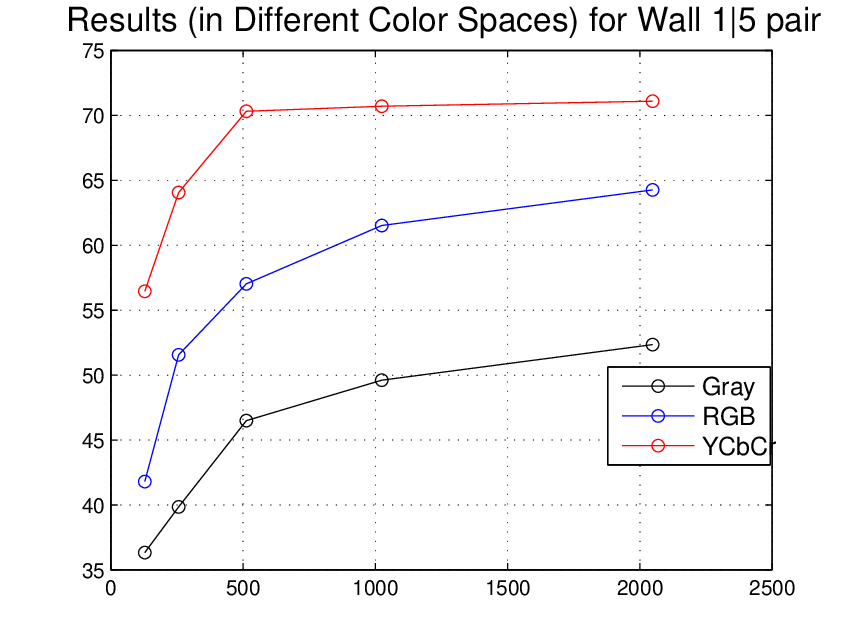}
	\end{center}
\caption{Performances of Gray, RGB and YCbCr Samplings vs. Number of Samples.}
\label{wall14_wall15}
\end{figure}

\begin{figure}[h!]
	\begin{center}
		\includegraphics [width=12.8cm, height=8cm] {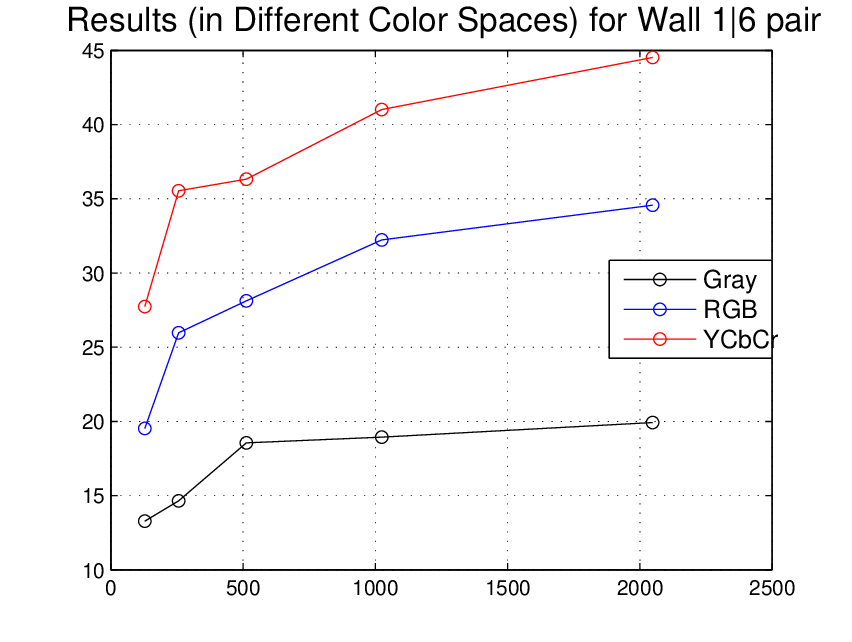}
		\includegraphics [width=12.8cm, height=8cm] {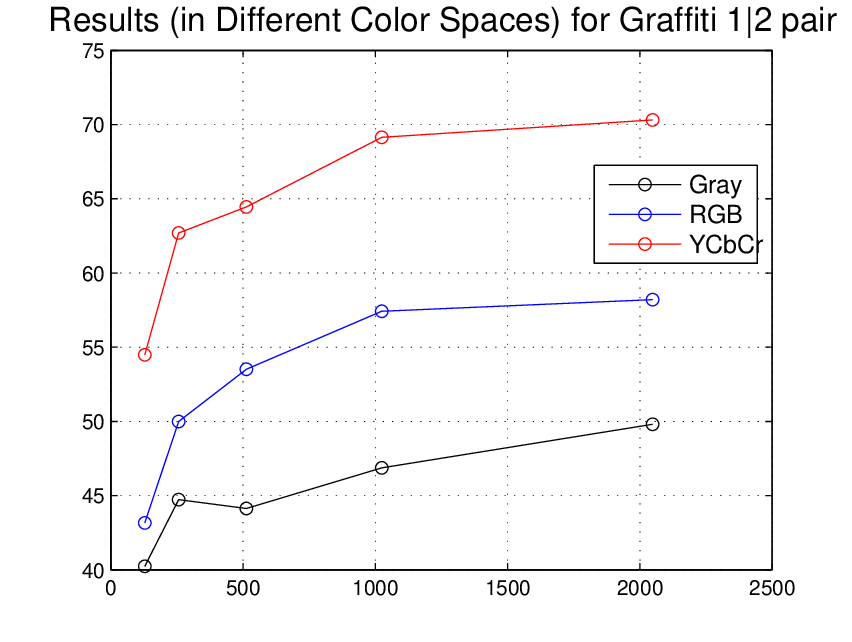}
	\end{center}
\caption{Performances of Gray, RGB and YCbCr Samplings vs. Number of Samples.}
\label{wall16}
\end{figure}

\begin{figure}[]
	\begin{center}
		\includegraphics [width=12.8cm, height=8cm] {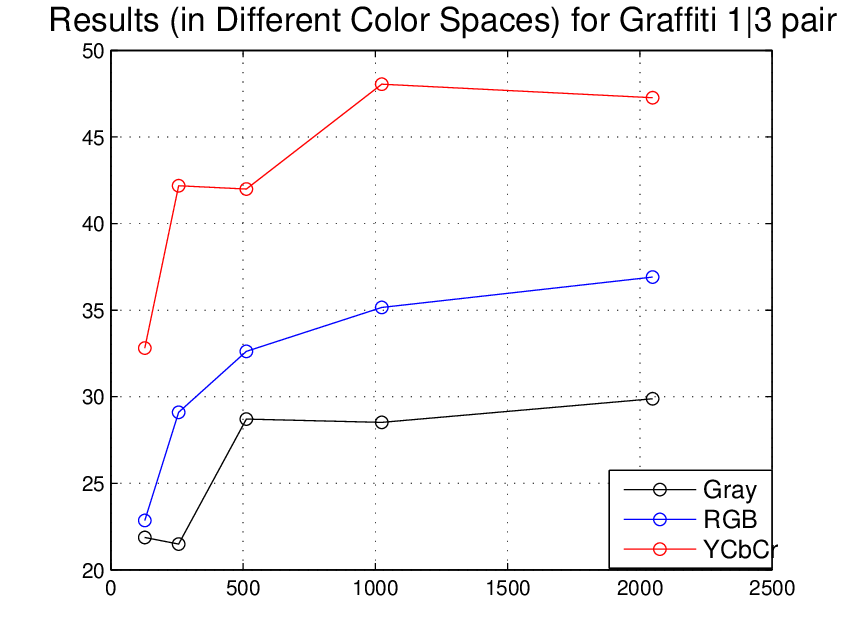}
		\includegraphics [width=12.8cm, height=8cm] {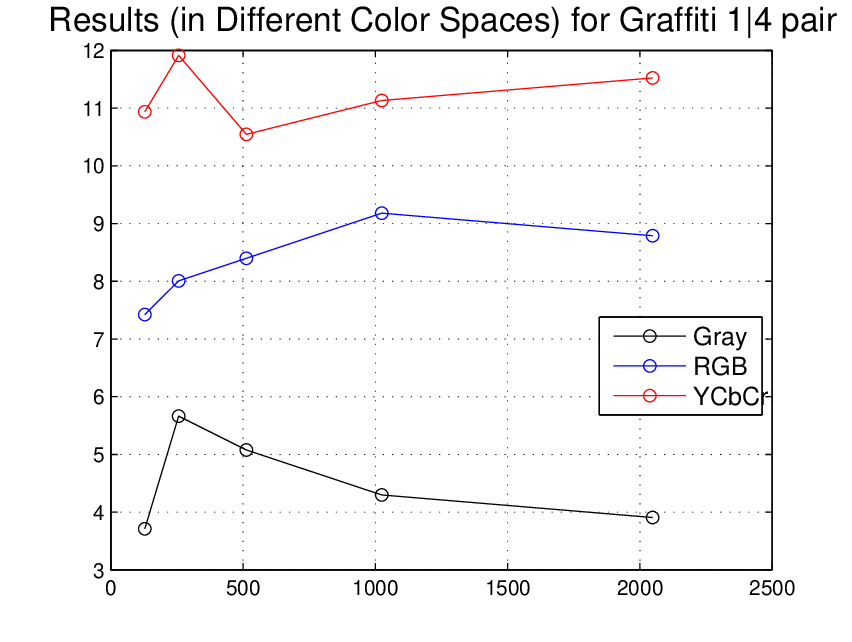}
	\end{center}
\caption{Performances of Gray, RGB and YCbCr Samplings vs. Number of Samples.}
\label{graf13_graf14}
\end{figure}

\begin{figure}[h!]
	\begin{center}
		\includegraphics [width=12.8cm, height=8cm] {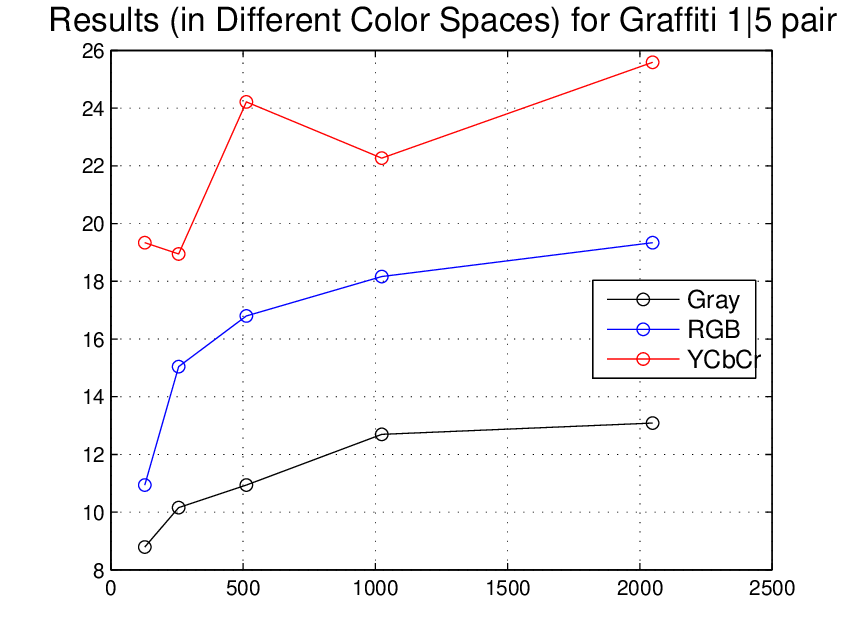}
		\includegraphics [width=12.8cm, height=8cm] {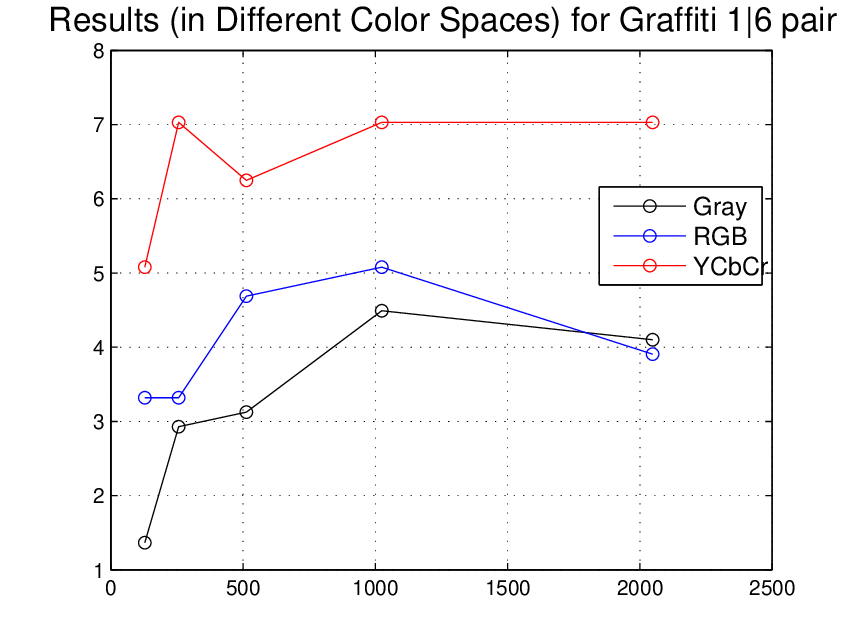}
	\end{center}
\caption{Performances of Gray, RGB and YCbCr Samplings vs. Number of Samples.}
\label{graf15_graf16}
\end{figure}

\begin{figure}[]
	\begin{center}
		\includegraphics [width=12.8cm, height=8cm] {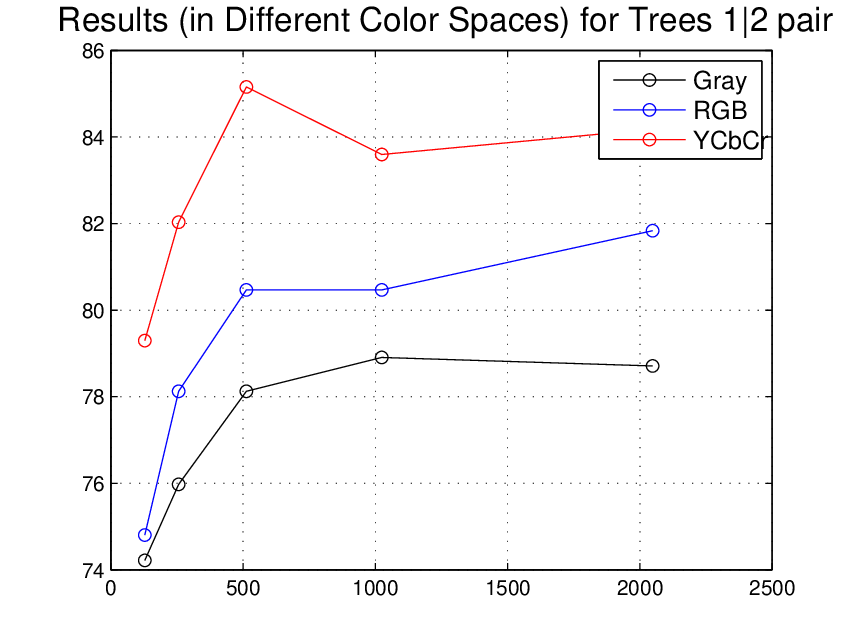}
		\includegraphics [width=12.8cm, height=8cm] {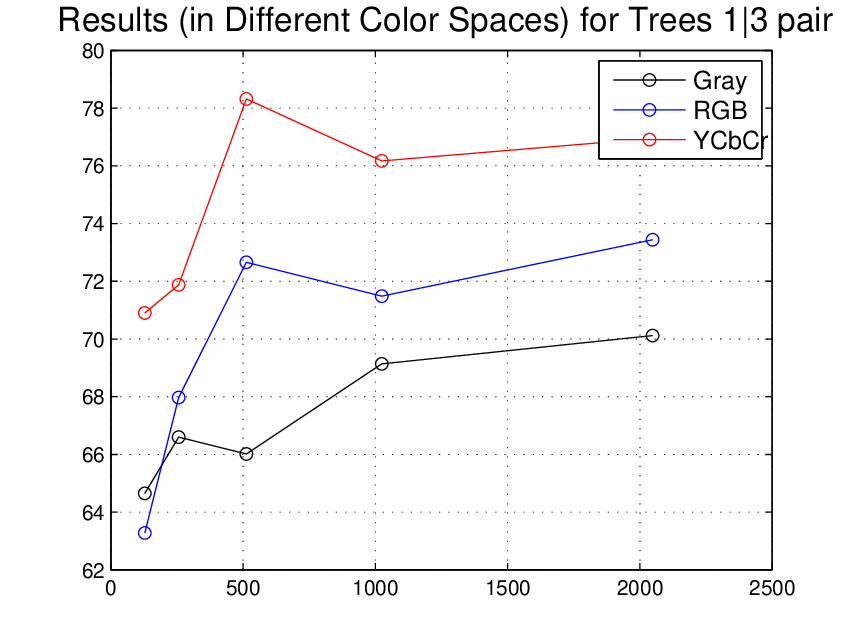}
	\end{center}
\caption{Performances of Gray, RGB and YCbCr Samplings vs. Number of Samples.}
\label{trees12_trees13}
\end{figure}

\begin{figure}[h!]
	\begin{center}
		\includegraphics [width=12.8cm, height=8cm] {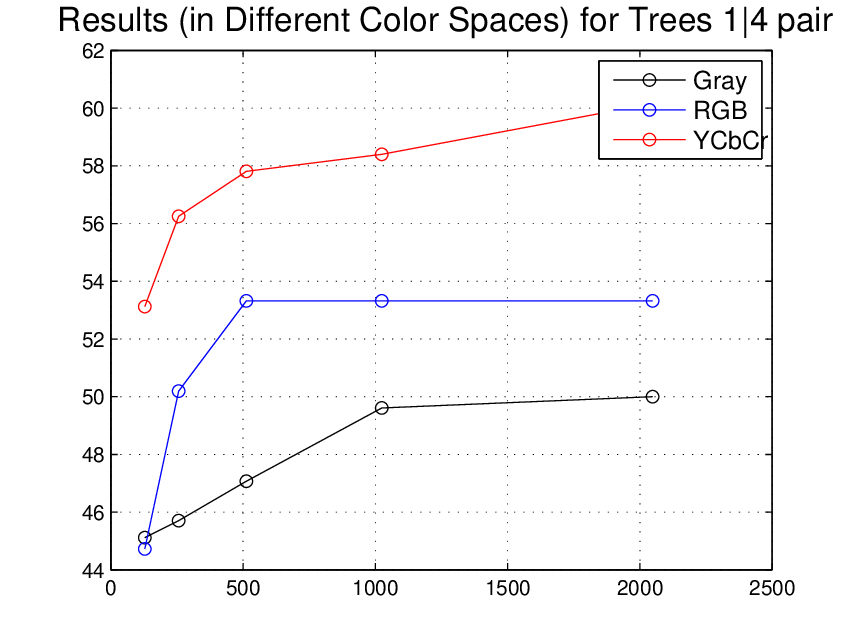}
		\includegraphics [width=12.8cm, height=8cm] {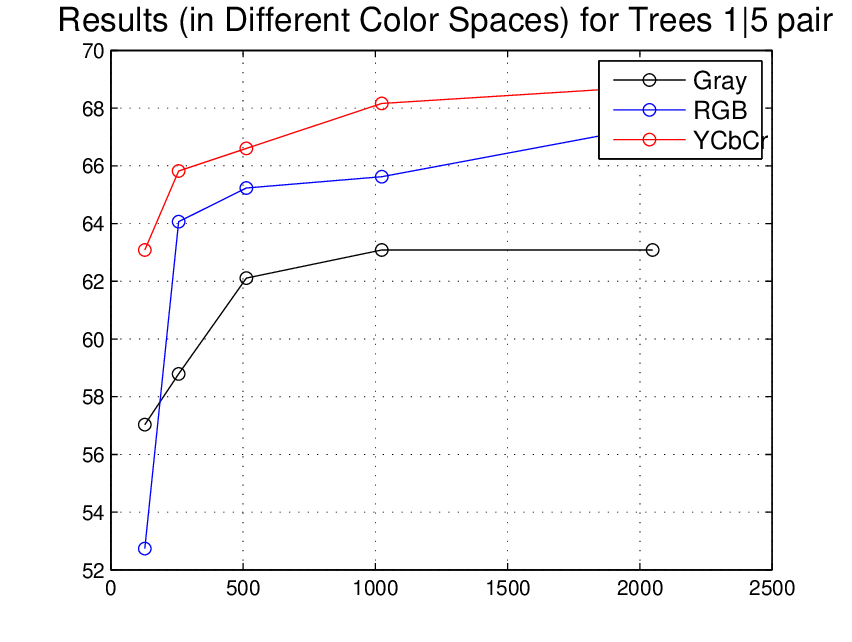}
	\end{center}
\caption{Performances of Gray, RGB and YCbCr Samplings vs. Number of Samples.}
\label{trees14_trees15}
\end{figure}

\begin{figure}[t]
	\begin{center}
		\includegraphics [width=12.8cm, height=8cm] {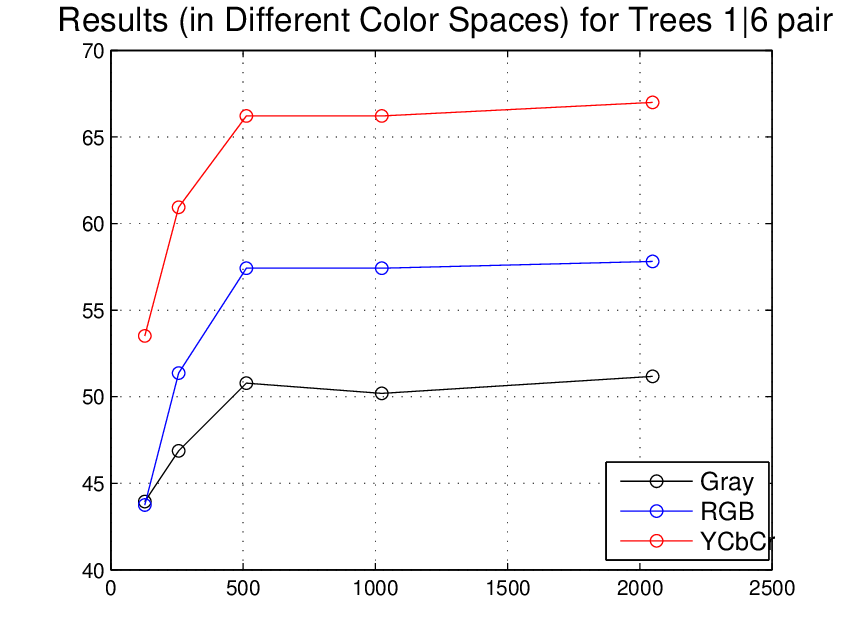}
		\includegraphics [width=12.8cm, height=8cm] {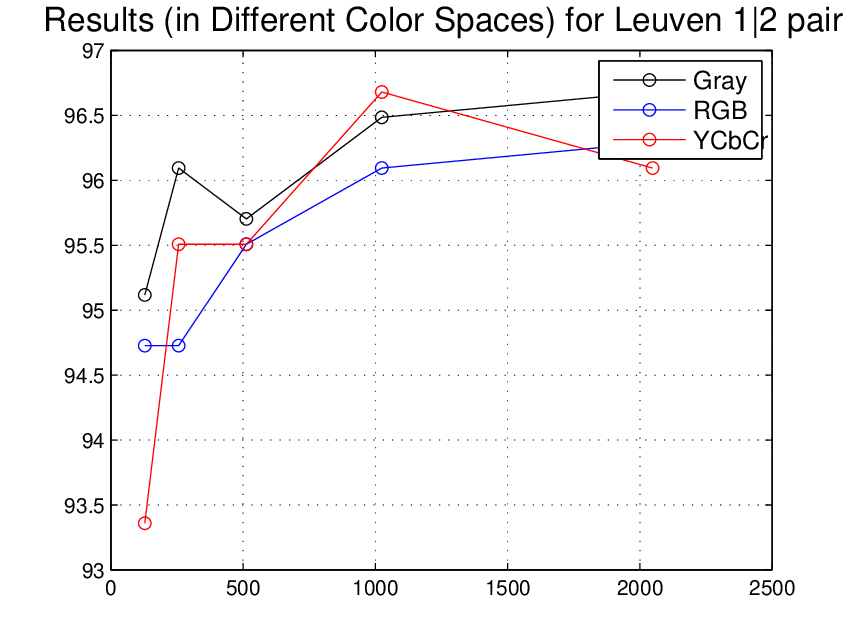}
	\end{center}
\caption{Performances of Gray, RGB and YCbCr Samplings vs. Number of Samples.}
\label{trees16}
\end{figure}

\begin{figure}[h!]
	\begin{center}
		\includegraphics [width=12.8cm, height=8cm] {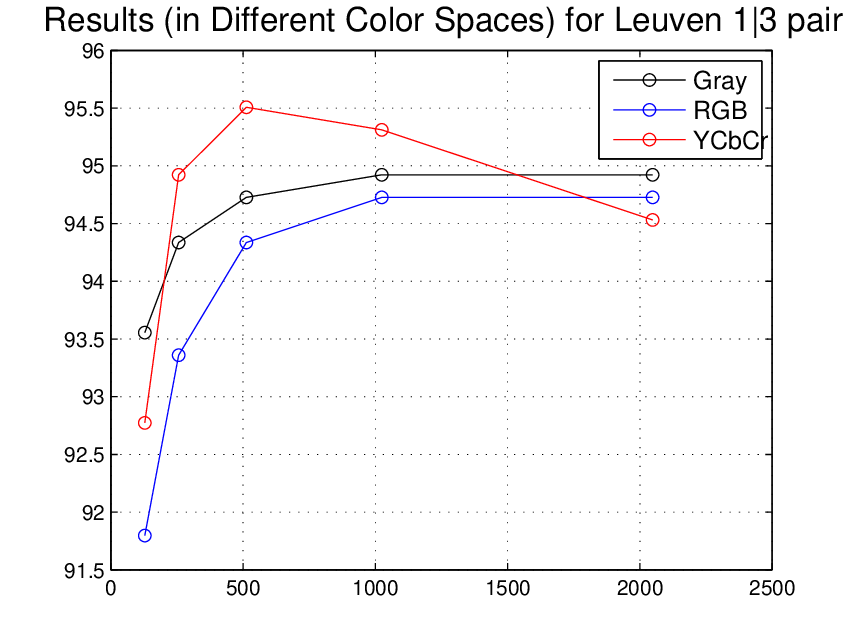}
		\includegraphics [width=12.8cm, height=8cm] {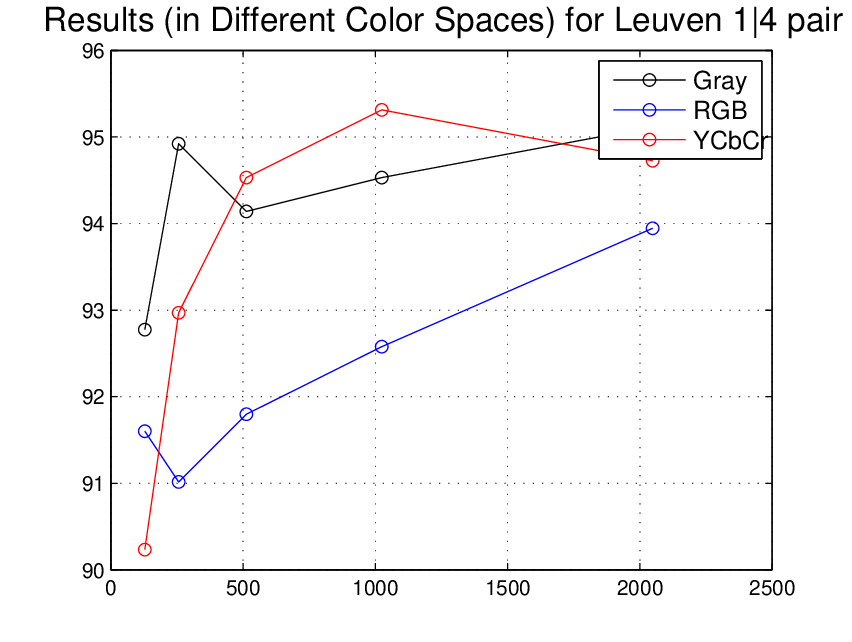}
	\end{center}
\caption{Performances of Gray, RGB and YCbCr Samplings vs. Number of Samples.}
\label{leuven13_leuven14}
\end{figure}

\begin{figure}[h!]
	\begin{center}
		\includegraphics [width=12.8cm, height=8cm] {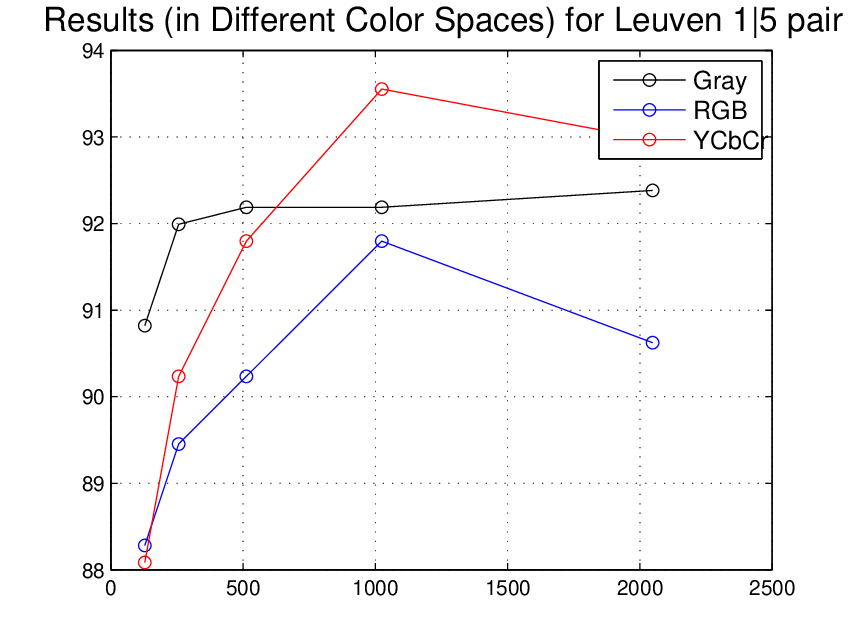}
		\includegraphics [width=12.8cm, height=8cm] {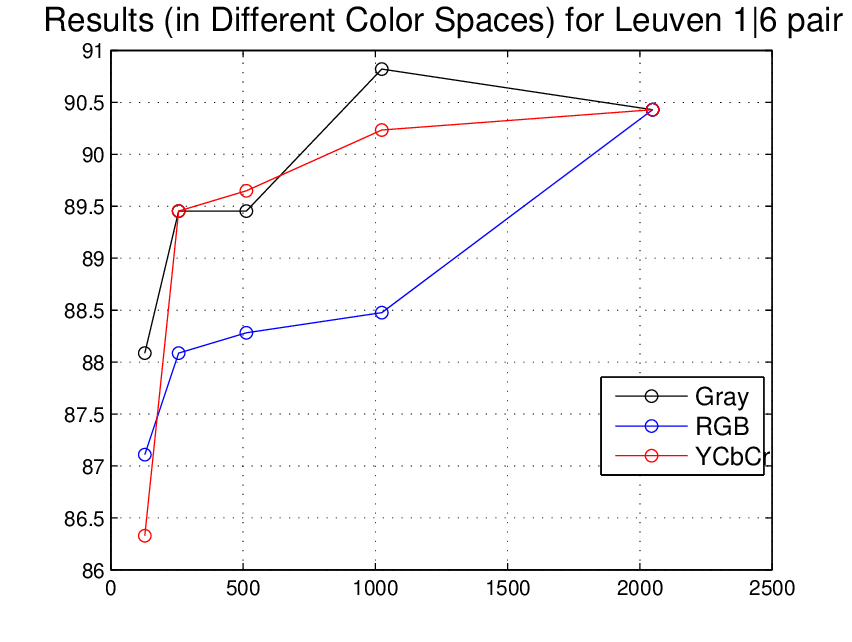}
	\end{center}
\caption{Performances of Gray, RGB and YCbCr Samplings vs. Number of Samples.}
\label{leuven15_leuven16}
\end{figure}

\begin{figure}[]
	\begin{center}
		\includegraphics [width=12.8cm, height=8cm] {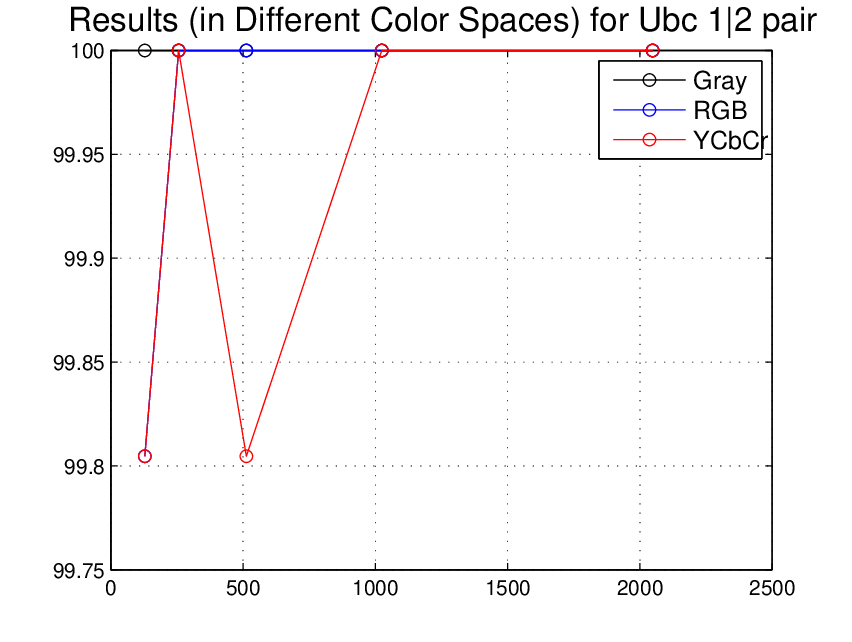}
		\includegraphics [width=12.8cm, height=8cm] {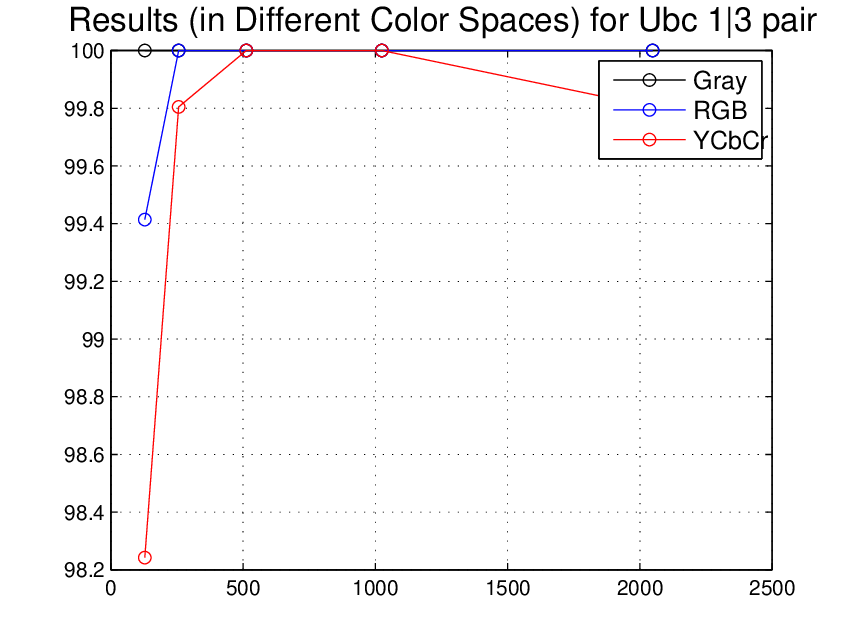}
	\end{center}
\caption{Performances of Gray, RGB and YCbCr Samplings vs. Number of Samples.}
\label{ubc12_ubc13}
\end{figure}

\begin{figure}[h!]
	\begin{center}
		\includegraphics [width=12.8cm, height=8cm] {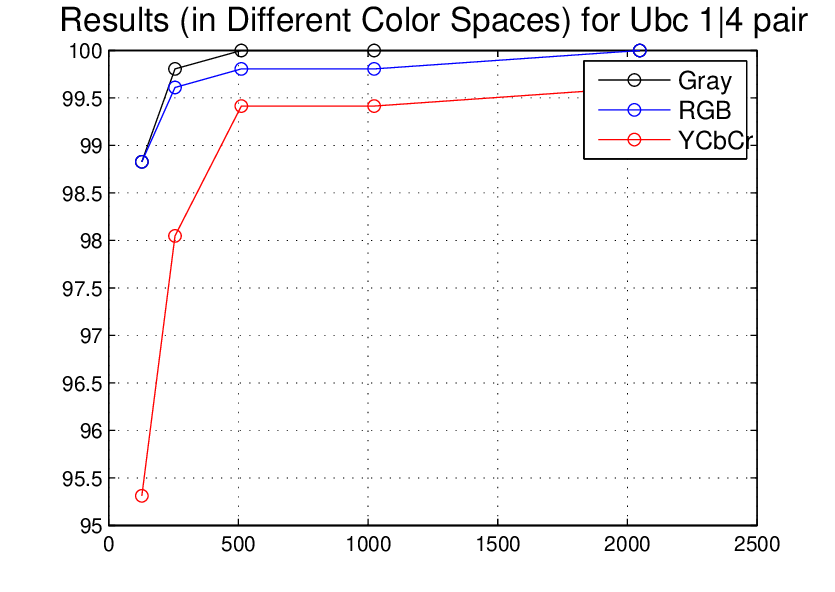}
		\includegraphics [width=12.8cm, height=8cm] {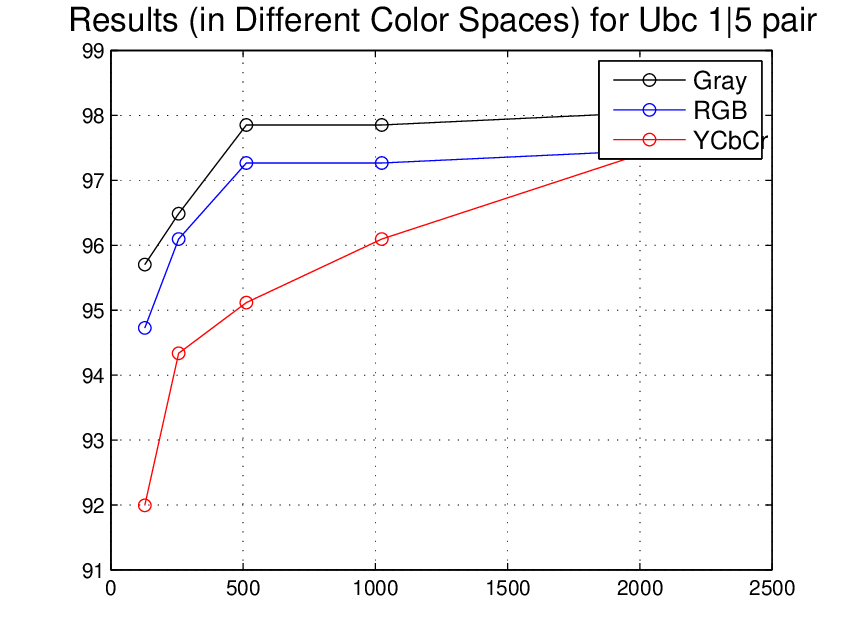}
	\end{center}
\caption{Performances of Gray, RGB and YCbCr Samplings vs. Number of Samples.}
\label{ubc14_ubc15}
\end{figure}

\begin{figure}[t]
	\begin{center}
		\includegraphics [width=12.8cm, height=8cm] {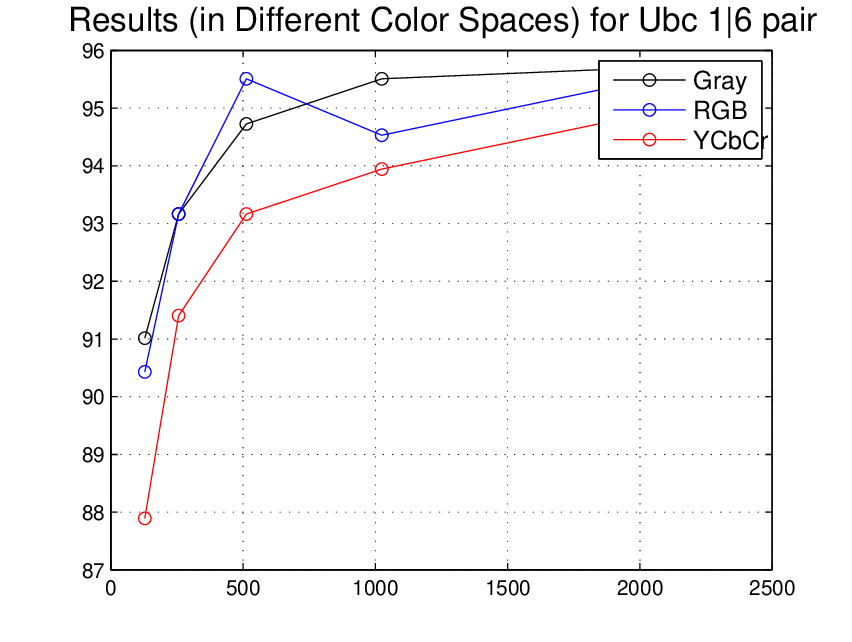}
	\end{center}
\caption{Performances of Gray, RGB and YCbCr Samplings vs. Number of Samples.}
\label{ubc16}
\end{figure}

Effect of the number of binary comparisons on Gray, RGB and YCbCr descriptors (BRIEF) are shown in \cref{wall12_wall13,wall14_wall15,wall16,graf13_graf14,graf15_graf16,trees12_trees13,trees14_trees15,trees16,leuven13_leuven14,leuven15_leuven16,ubc12_ubc13,ubc14_ubc15,ubc16}.
As shown in these figures, RGB-BRIEF and YCbCr-BRIEF descriptor performance continue to increase with the number of tests. 

\section{Effects of the patch sizes}
\begin{figure}[]
\begin{center}
	\subfloat[][] {\includegraphics [width=6cm, height=4cm] {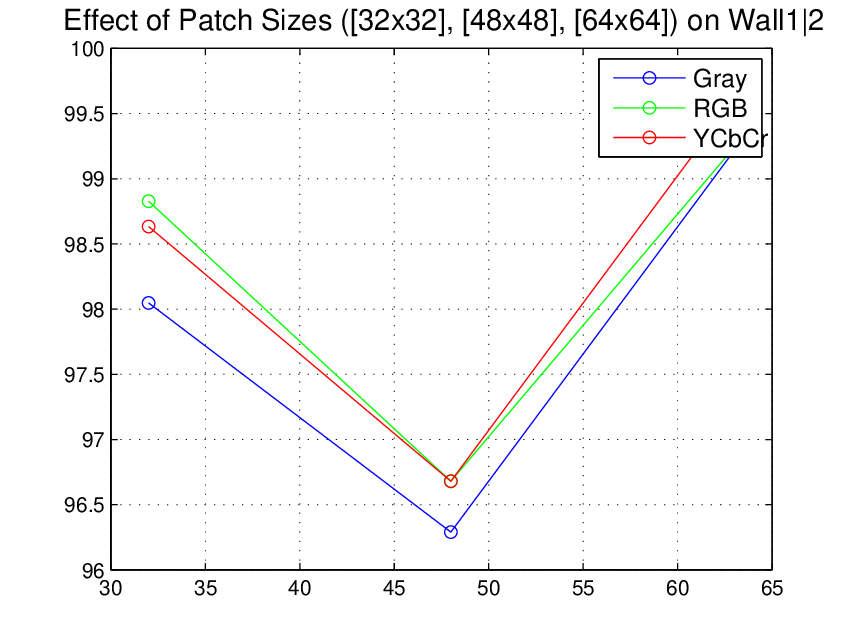}}
	\subfloat[][] {\includegraphics [width=6cm, height=4cm] {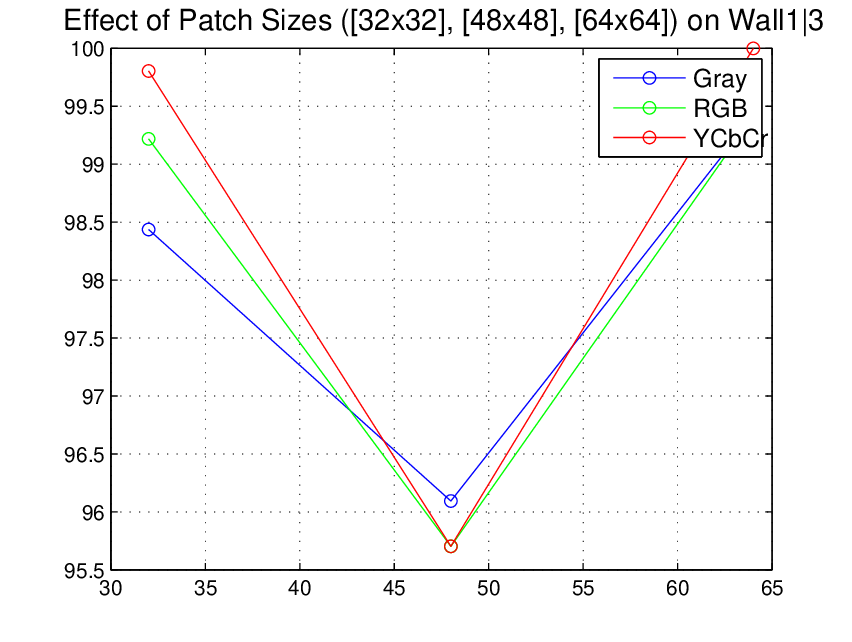}}
	\quad
	\subfloat[][] {\includegraphics [width=6cm, height=4cm] {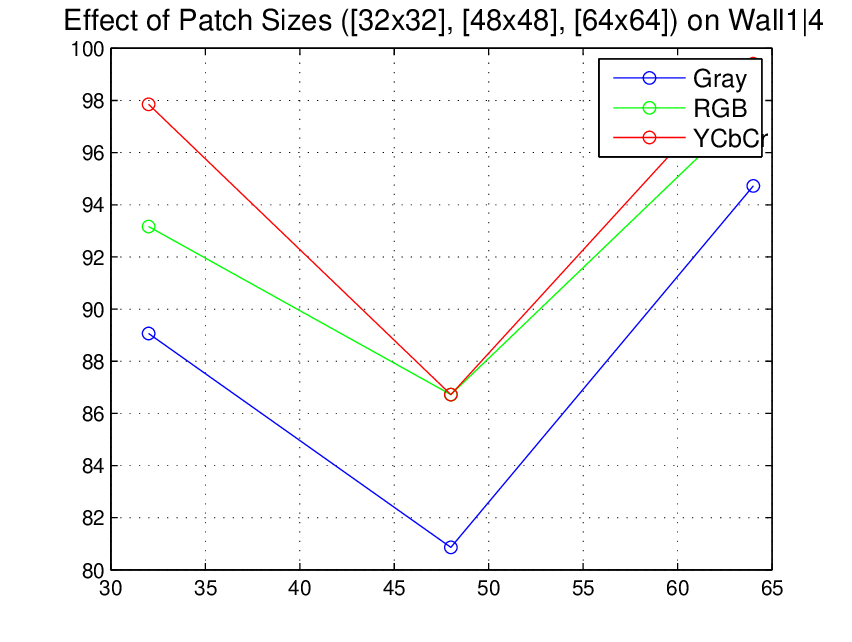}}
	\subfloat[][] {\includegraphics [width=6cm, height=4cm] {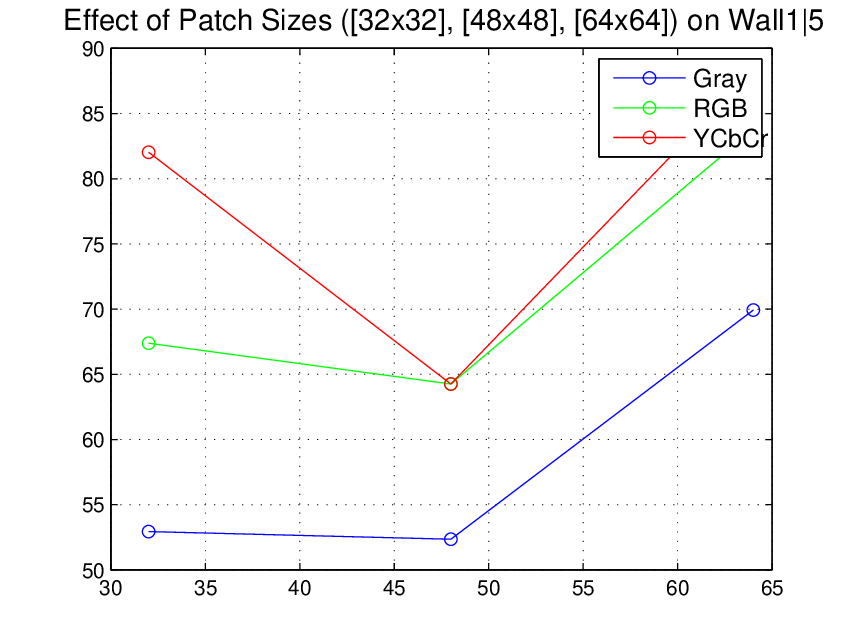}}
	\quad
	\subfloat[][] {\includegraphics [width=6cm, height=4cm] {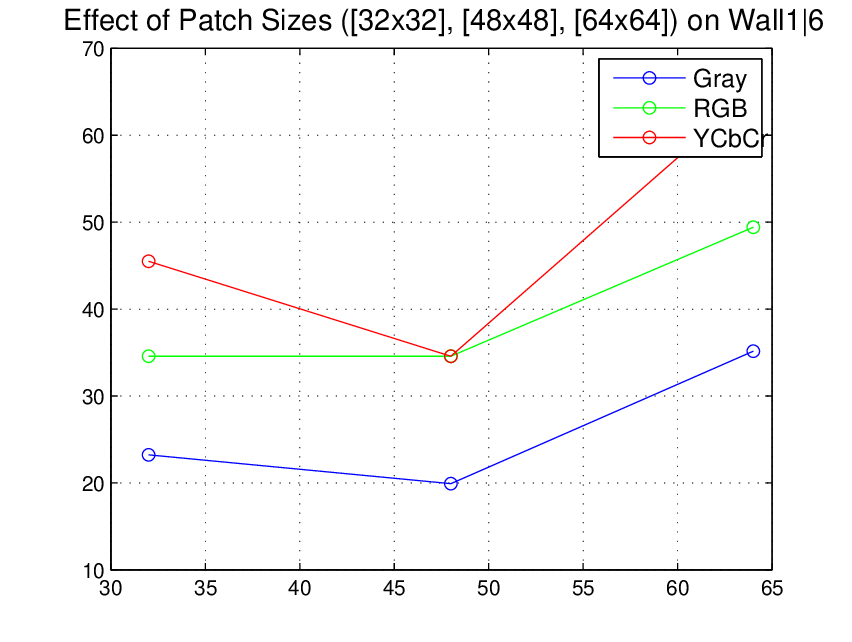}}

\caption{Performances of Gray, RGB and YCbCr Samplings vs. Number of Samples.}
\label{wall12_patch_size}
	\end{center}
\end{figure}

In order to examine the consistency of our color descriptors, we looked for the effect of patch sizes. If all descriptors behave in a similar way, then it would strengthen the stability of proposed descriptors and it was as thought. All Gray, RGB and YCbCr descriptors (BRIEF) behave in a similar manner for each patch size as shown in \ref{wall12_patch_size}.

%

